\pdfoutput=1
\documentclass[lettersize,journal]{IEEEtran}
\usepackage{amsmath,amsfonts}
\usepackage{algorithmic}
\usepackage{algorithm}
\usepackage{array}
\usepackage[caption=false,font=normalsize,labelfont=sf,textfont=sf]{subfig}
\usepackage{textcomp}
\usepackage{stfloats}
\usepackage{url}
\usepackage{verbatim}
\usepackage{graphicx}
\usepackage{cite}
\usepackage{multirow}
\usepackage{placeins}
\usepackage[hidelinks]{hyperref}

\begin{document}

\title{RAIBO-Wagon: Real-Time Navigation System for a Legged Robot Towing a Wagon}

\author{Insung Yang, Eunchae Kim, Jungwoo Hur, Seunghun Jeon, Donghyuk Choi, and Jemin Hwangbo$^{*}$%
\thanks{All authors are with the Robotics and Artificial Intelligence Lab, KAIST, Daejeon, South Korea.}%
\thanks{$^{*}$Corresponding author: jhwangbo@kaist.ac.kr}%
}

\markboth{JOURNAL SUBMISSION}{JOURNAL SUBMISSION}

\maketitle

\begin{abstract}
Legged robots have demonstrated advanced capabilities across a wide range of real-world tasks, yet much of this progress concerns the robot controlling only itself. Once the robot is coupled to an external object---particularly an underactuated one---the dynamics of the combined assembly become more complex. Towing a wagon is a common practical scenario, where coupled dynamics make reliable navigation challenging. This difficulty is amplified in confined spaces, where collision-free navigation requires finding a narrow set of feasible paths under strict real-time constraints. To address this, we propose a navigation framework that combines Conditional Flow Matching (CFM) with Model Predictive Path Integral (MPPI) control over a learned Forward Dynamics Model (FDM). The CFM model mitigates the initialization problem of MPPI and accelerates its convergence, while MPPI with the FDM recovers feasible solutions beyond the CFM model's training distribution by iterative refinement. We integrate the framework into a full-stack navigation system for RAIBO-Wagon, a quadruped towing a custom wagon, and validate it in real-world outdoor experiments on 15\,m-long courses through confined spaces, with all sensing and computation performed fully onboard. The system achieves an 85\% success rate across 40 trials, towing a total load of up to 40\,kg---nearly 100\% of RAIBO's body weight. 

Project page: \url{https://insungyang.github.io/raibo-wagon/}

\end{abstract}


\section{Introduction}

\IEEEPARstart{Q}{uadrupedal} robots have demonstrated advanced locomotion capabilities over challenging terrain~\cite{hwangbo2019learning, lee2020learning, ji2022concurrent, choi2023learning, kim2025high, lee2026quadruped}. Recently, they have been applied to tasks beyond locomotion, such as loco-manipulation~\cite{jung2026learning, he2026pake, liu2024visual}, whole-body manipulation~\cite{jeon2023learning, yuan2025caiman}, and pedipulation~\cite{arm2024pedipulate, stolle2024perceptive}. Payload transport can further extend their practical utility, but the load a quadruped can carry on its own body is limited. Towing a wagon overcomes this limitation, combining the load capacity of a wheeled carrier with the terrain traversability of legged locomotion.

However, coupling the quadruped and the wagon yields an underactuated system with an enlarged footprint, and the wagon's wheels further restrict its motion to its heading. These properties make it difficult to maneuver the wagon through constrained environments such as narrow corridors and doorways. To address this issue, prior work has relied on model-based trajectory optimization using explicit quadruped-wagon dynamics modeling, which incurs high computational costs~\cite{zhang2025hybrid}. As a consequence, this approach has been demonstrated only in short-horizon settings or with preplanned trajectories, precluding its application to real-time adaptive execution. These limitations motivate the objective of this work: a navigation framework for a quadruped towing a wagon that generates dynamically feasible commands under real-time constraints in confined spaces.

\begin{figure}[t]
    \centering
    \includegraphics[width=1.0\linewidth]{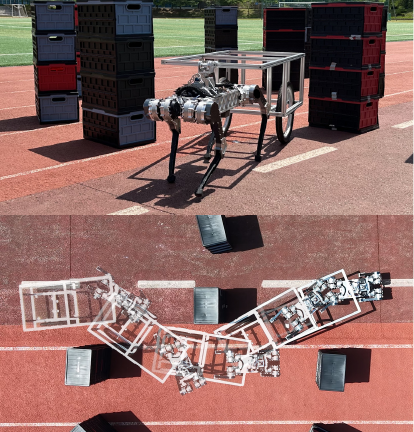}
    \caption{This figure shows snapshots from real-world deployment experiments of RAIBO towing a wagon during navigation. The top image shows a front view of the robot navigating while pulling the wagon, and the bottom image provides a drone perspective of the same scenario.}
    \label{fig:intro}
\end{figure}

This objective calls for an approach that is both dynamics-aware and computationally lightweight. One promising direction combines learning-based dynamics modeling with sampling-based optimization. In particular, Model Predictive Path Integral (MPPI) control has proven effective on wheeled robots~\cite{williams2016aggressive, williams2017information}: it rolls out candidate trajectories through a learned forward dynamics model (FDM) and optimizes actions without gradient computation. Unlike policies based purely on imitation learning or reinforcement learning (RL), this formulation reasons over the dynamics at inference time, yet avoids the computational cost of full model-based trajectory optimization. Recent work has extended this approach to quadrupedal navigation, demonstrating its effectiveness on legged platforms~\cite{kim2022learning, roth2025learned}. In our setting, however, standard warm-starting in MPPI, such as time-shifting the previous action sequence, struggles to find valid trajectories within a real-time budget. Coupling the quadruped and the wagon confines feasible trajectories to a narrow region of the command space, which random perturbations around the shifted sequence rarely reach.

To this end, we introduce a navigation framework in which MPPI is warm-started by a Conditional Flow Matching (CFM) model~\cite{lipman2022flow}. CFM models are well suited for real-time warm-starting, as they generate promising sample candidates in only a few integration steps. To train the CFM model on high-quality demonstrations, we construct an offline expert pipeline free from real-time constraints: it first plans a path using a learned collision-risk predictor that evaluates the full quadruped-wagon configuration, and then tracks this path with MPPI. Distilling the resulting commands into the CFM model embeds this costly wagon-aware planning into a single generative model, which supplies MPPI with feasible candidates at runtime. Deployed online, the CFM model and MPPI are complementary: the CFM model mitigates MPPI's initialization problem by concentrating samples in high-reward regions. In addition, the online MPPI recovers feasible solutions beyond the CFM model's learned distribution by re-optimizing the samples. Together, these properties enable real-time navigation of a quadruped towing a wagon in confined spaces.

We validate our framework through experiments in both simulation and the real world. In simulation, our CFM-based warm-start outperforms vanilla MPPI, more than doubling the success rate in the most constrained settings. The CFM warm-start also outperforms alternative generative architectures, and the full framework surpasses alternative ways of combining the CFM model with the FDM. For real-world deployment, we use RAIBO2, a quadrupedal robot~\cite{lee2026quadruped}, which pulls a custom wagon that is attached to the robot through a 3-DOF passive joint with an encoder measuring the relative yaw angle. By relying solely on onboard LiDAR for mapping and localization with no external infrastructure, it navigates a 15\,m outdoor course as shown in Fig.~\ref{fig:intro}. Across 40 trials on eight different terrains with both static obstacles and pedestrians, it achieves an 85\% success rate. Throughout these trials, it tows a total load of up to 40\,kg including the wagon, nearly equal to its own body weight.

In summary, the main contributions of this work are as follows:
\begin{enumerate}
    \item We present a full-stack quadrupedal wagon-towing system, integrating a wagon-aware locomotion controller with a navigation framework that operates purely on onboard sensing and computation without preplanned trajectories.

   \item We achieve real-time sampling-based navigation under coupled quadruped-wagon dynamics by warm-starting MPPI with an offline-distilled CFM model and rolling out candidates through a learned forward dynamics model.

   \item We demonstrate towing a total load of up to 40\,kg, nearly equal to the robot's own body weight, and validate the navigation system through extensive real-world experiments.
\end{enumerate}

\section{Related Work}
\subsection{Legged Robot Navigation}
 
\subsubsection{Path Planning}
Path planning tackles navigation by producing a route for the robot to follow toward its goal. One strategy is to learn traversability or motion costs, biasing the route toward terrain that a quadruped can negotiate easily~\cite{wellhausen2021rough, wellhausen2023artplanner, yang2021real}. Rather than learning these costs, TRG-Planner~\cite{lee2025trg} builds a traversal risk graph directly from the terrain map, scaling terrain-aware planning to large environments. Other works instead plan directly from onboard perception: iPlanner~\cite{yang2023iplanner} trains a planner via imperative learning; ViPlanner~\cite{roth2024viplanner} in turn extends it with semantic terrain understanding for zero-shot sim-to-real transfer. Wild Visual Navigation~\cite{mattamala2025wild} learns traversability online in a self-supervised manner from pre-trained visual features, adapting to previously unseen terrain within minutes of in-field operation.

These planners produce a geometric route under the assumption that the quadruped can track it closely. For our task, however, this assumption breaks down: since the wagon is underactuated, a geometrically valid path may be dynamically infeasible. Therefore, path planning serves only to supply a coarse global route, while feasibility should be resolved downstream in command space.

\subsubsection{Reinforcement Learning for Local Navigation}
To reason directly over velocity commands rather than geometric paths, many works train navigation policies with reinforcement learning (RL). For instance, an RL-based navigation policy achieves kilometer-scale urban navigation on wheeled-legged robots~\cite{lee2024learning}. Similarly, a perception-resilient end-to-end RL policy withstands sensor degradation and failure~\cite{zhang2024resilient}. In a related effort, a spatially enhanced recurrent memory supports long-range mapless navigation~\cite{yang2025spatially}. More closely related to our setting, a path-conditioned local planner uses a reference path as contextual guidance for an RL policy, enabling reactive local navigation~\cite{haro2026path}.

However, a learned value function used in RL training is estimated by bootstrapping, so its error can accumulate over long horizons. A policy trained against this value function inherits the same error. This limitation is critical in our setting, where the coupling of the quadruped and the wagon leaves little margin for dynamically infeasible actions. Instead, sampling-based optimization can mitigate this limitation by explicitly rolling out candidate actions through a dynamics model at inference time.

\subsubsection{Forward Dynamics Models with MPPI}
One of the most widely used sampling-based optimization methods is MPPI. On wheeled platforms, MPPI has been widely used and has demonstrated high-speed off-road driving~\cite{williams2017information, williams2016aggressive, drews2017aggressive, williams2018robust}. Several works also incorporate raw images directly into this framework~\cite{kahn2021badgr, lee2024crop}.

In the legged robotics field, Kim~\textit{et al.}~\cite{kim2022learning} first demonstrated learned FDM-based MPPI for safe quadruped navigation, and Roth~\textit{et al.}~\cite{roth2025learned} extended this with a 3D perceptive FDM. However, the additional complexity introduced by the wagon makes the optimization harder. Therefore, MPPI requires more samples and iterations to reach a feasible solution, a cost that is incompatible with real-time operation. This limitation motivates our use of a generative model as a warm-start: the CFM model provides a feasible trajectory, which accelerates convergence.

\subsection{Generative Models for Robot Planning}

\subsubsection{Imitation Learning with Generative Models}
Generative models have recently demonstrated strong performance in robot control by learning the multimodal distribution of expert behavior. Diffusion Policy~\cite{chi2025diffusion} demonstrated that a denoising diffusion model trained on expert demonstrations can capture the contact-rich structure of manipulation tasks, outperforming deterministic behavioral cloning by modeling the full conditional action distribution. In place of diffusion, flow matching~\cite{lipman2022flow}, which learns a vector field with nearly straight transport paths and thus generates samples in only a few integration steps, has been increasingly adopted as the generative backbone of vision-language-action models~\cite{black2024pi_0, bjorck2025gr00t, shukor2025smolvla}. Beyond manipulation, generative policies have spread to other robot domains. In locomotion, DynaFlow~\cite{lee2025dynaflow} applied conditional flow matching to quadrupedal motion generation, embedding a differentiable dynamics model into the flow matching framework. For navigation, NoMaD~\cite{sridhar2024nomad} proposed a goal-masked diffusion policy trained on diverse robot platforms, and LDP~\cite{yu2024ldp} demonstrated a local diffusion planner for collision avoidance deployed on real mobile robots. While these methods effectively capture multimodal behavior, they are trained purely by imitation and do not incorporate explicit task or safety signals at inference time, leaving the generated commands potentially unsafe in novel environments.

\subsubsection{Generative Models as Trajectory Planners}
A parallel line of work uses generative models as trajectory planners, incorporating task costs directly into the generative process to bias samples toward task-feasible trajectories. Diffuser~\cite{janner2022planning} treats the full state-action trajectory as the generation target and steers the denoising process with reward gradients. In robot motion planning, MPD~\cite{carvalho2023motion} learns a trajectory prior and refines it via cost-guided posterior sampling. In navigation, NaviDiffusor~\cite{zeng2025navidiffusor} applies cost-guided diffusion to visual navigation for zero-shot transfer to novel environments, and DTG~\cite{liang2024dtg} conditions a diffusion model on LiDAR and target information for large-scale mapless navigation. A related strategy treats the generative output as a candidate pool, selecting or iteratively refining the highest-reward samples with a learned critic or simulator, as in Diffusion ES~\cite{yang2024diffusion}. Such guidance- and selection-based planners bias sampling toward task-feasible regions, yet their outputs remain tied to the learned distribution: selection can only choose among generated samples, and cost guidance mainly steers samples within or near the learned modes.

A related line of work instead couples generative models with sampling-based optimization, pushing solutions beyond the training distribution through iterative refinement. Mizuta~\textit{et al.}~\cite{mizuta2025unified} run this coupling online, with flow-matching samples serving as informed priors for MPPI and the refined solution warm-starting the next generation. Generative Predictive Control~\cite{kurtz2025generative} distills the solutions of a sampling-based predictive controller into a flow-matching policy through iterative bootstrapping, and a concurrent approach~\cite{brudermuller2026generative} carries this coupling to runtime, training a flow-matching model on CEM solutions and injecting its samples back into the optimizer. These methods were developed for settings with a broad feasible set, where many trajectories are acceptable. The quadruped-wagon system, however, admits only a narrow set of feasible trajectories. Finding such trajectories with a sampling-based method requires both a large sample count and several refinement iterations, which in turn demand a dynamics model that is both fast and accurate. A simplified dynamics model~\cite{mizuta2025unified} evaluates samples quickly but cannot fully capture the robot dynamics, whereas a full physics simulator~\cite{kurtz2025generative, brudermuller2026generative} is accurate but too slow to complete the required sampling within the real-time budget. Our framework resolves this trade-off with a learned forward dynamics model that captures the coupled quadruped-wagon dynamics while remaining fast enough to support large-sample, multi-iteration search in real time.

\subsection{Tractor-Trailer Navigation}

Tractor-trailer navigation has been extensively studied in the wheeled vehicle domain. Ljungqvist~\textit{et al.}~\cite{ljungqvist2019path} proposed an integrated path planning and path-following control framework for a general 2-trailer with a car-like tractor. Chen~\textit{et al.}~\cite{chen2025universal} proposed a universal navigation system for vehicle-trailer platforms that combines neural kinematics with online residual learning, enabling adaptation to diverse trailer configurations and payloads without manual re-tuning. Tracailer~\cite{xu2025tracailer} presented a time-efficient trajectory planner for tractor-trailer robots in unstructured environments, achieving real-time planning while explicitly accounting for the geometric constraints of the coupled system. Despite this extensive body of work in wheeled robotics, these methods cannot be directly deployed on legged platforms. They rely on a rigid-body kinematic model in which the tractor is treated as a smoothly steerable vehicle, whereas a legged robot exhibits dynamics that such a model cannot capture.

In the legged robot domain, Zhang~\textit{et al.}~\cite{zhang2025hybrid} attached a cable-towed trailer to a quadruped and developed a hybrid dynamics model capturing the cable's taut and slack states for trajectory planning with mode transitions. However, this approach plans trajectories offline and only tracks them online, so it does not support real-time navigation, limiting its applicability to real-world deployment. To the best of our knowledge, no prior work has addressed real-time tractor-trailer navigation for a quadruped with a wagon in outdoor environments.

\section{Preliminaries}
\subsection{Model Predictive Path Integral (MPPI)}
\label{sec:preliminary_mppi}
Model Predictive Path Integral (MPPI) is a sampling-based control method that offers a simple yet effective approach to selecting an action sequence for tracking a given path. Given the current observation $o_t$ and a reference path $\mathcal{P}$, MPPI seeks the action sequence $\bar{\mathbf{a}} = (a_0, \ldots, a_{T-1})$ over a horizon $T$ that maximizes the expected cumulative reward:
\begin{equation}
  \bar{\mathbf{a}}^* = \arg\max_{\bar{\mathbf{a}}} \;
  \mathbb{E}\bigl[ R(\tilde{\mathbf{s}}, \mathcal{P}) \bigr],
\end{equation}
where $\tilde{\mathbf{s}} = (\tilde{s}_1, \ldots, \tilde{s}_T)$ is the rollout
predicted from the action sequence in a single forward pass,
$\tilde{\mathbf{s}} = f(\bar{\mathbf{a}}, o_t)$. Rather than optimizing this
directly, MPPI draws $K$ candidates by perturbing the previous solution with
decaying Gaussian noise,
\begin{equation}
  \delta \mathbf{a}^k_{(i)} \sim \mathcal{N}\!\left(\mathbf{0},\;
  \alpha^i \boldsymbol{\Sigma}\right), \quad \alpha \in (0, 1),
\end{equation}
where $\boldsymbol{\Sigma}$ is a fixed covariance and the decay $\alpha$ encourages
broad exploration early and fine refinement as optimization converges. Each
candidate action sequence $\mathbf{a}^k_{(i)} = \bar{\mathbf{a}}_{(i)} +
\delta\mathbf{a}^k_{(i)}$ is scored by $R^k$, and the solution is updated by a softmax-weighted sum:
\begin{equation}
  \bar{\mathbf{a}}_{(i+1)} \leftarrow \bar{\mathbf{a}}_{(i)} +
  \sum_{k=1}^{K} w^k \, \delta \mathbf{a}^k_{(i)},
\end{equation}
\begin{equation}
  w^k = \frac{\exp\!\left(\gamma(R^k - R_{\max})\right)}
             {\sum_{l=1}^{K} \exp\!\left(\gamma(R^l - R_{\max})\right)},
\end{equation}
where $\gamma > 0$ is an inverse temperature and $R_{\max} = \max_k R^k$ ensures numerical
stability. This procedure requires no gradient through the dynamics, making it
well suited for learned forward models.

\subsection{Conditional Flow Matching with Classifier-Free Guidance}
\label{sec:cfg_training}

Flow Matching learns a time-dependent vector field
$v_\phi : \mathbb{R}^d \times [0,1] \rightarrow \mathbb{R}^d$
that transforms an initial distribution $p_0 = \mathcal{N}(\mathbf{0}, \mathbf{I})$
into a target data distribution $p_1$ via an ODE:
\begin{equation}
    \frac{d\mathbf{x}_t}{dt} = v_\phi(\mathbf{x}_t, t),
    \quad \mathbf{x}_0 \sim p_0.
\end{equation}
However, directly regressing on the marginal vector field is intractable,
as it requires integrating over all data.
Conditional Flow Matching (CFM)~\cite{lipman2022flow} resolves this by
defining a conditional probability path $p_t(\mathbf{x}|\mathbf{x}_1)$
given a data sample $\mathbf{x}_1 \sim p_1$, which has a closed-form
regression target. Using a Gaussian path with a small constant
$\sigma \ll 1$ that sets the minimum noise level,
\begin{equation}
    p_t(\mathbf{x} | \mathbf{x}_1) = \mathcal{N}\!\left(\mathbf{x} \mid
    t\,\mathbf{x}_1,\,(1-(1-\sigma)t)^2\mathbf{I}\right),
\end{equation}
the corresponding conditional vector field is
\begin{equation}
    u_t(\mathbf{x}_t | \mathbf{x}_1) =
    \frac{\mathbf{x}_1 - (1-\sigma)\mathbf{x}_t}{1-(1-\sigma)t},
\end{equation}
and the CFM training objective becomes
\begin{equation}
    \mathcal{L}_{\mathrm{CFM}}(\phi) =
    \mathbb{E}_{t,\,\mathbf{x}_1,\,\mathbf{x}_t}\!
    \left\| v_\phi(\mathbf{x}_t, t, c)
    - u_t(\mathbf{x}_t | \mathbf{x}_1) \right\|^2,
\end{equation}
where $c$ denotes an optional conditioning variable that the network takes as an
additional input. This objective is simulation-free and has been shown to yield
the same gradient as the marginal Flow Matching loss~\cite{lipman2022flow}. In practice, we take $\sigma \rightarrow 0$, which reduces the path to the straight interpolation used in Algorithm~\ref{alg:cfm_training}.

To strengthen adherence to the condition $c$ at inference,
we apply Classifier-Free Guidance (CFG)~\cite{ho2022classifier}.
During training, the condition $c$ is randomly dropped
with probability $p_{\mathrm{drop}}$, causing the model to learn
both conditional and unconditional vector fields simultaneously.
At inference, the two are interpolated with a guidance scale $w \geq 1$:
\begin{equation}
    \tilde{v}_\phi(\mathbf{x}_t, t, c) =
    w \cdot v_\phi(\mathbf{x}_t, t, c)
    + (1-w) \cdot v_\phi(\mathbf{x}_t, t, \emptyset),
\end{equation}
where $\emptyset$ denotes the null condition.
A sample is then generated by integrating $\tilde{v}_\phi$ from $t=0$ to $t=1$,
approximated via an Euler solver over a fixed number of steps.

\begin{figure*}[t]
    \centering
    \includegraphics[width=0.95\linewidth]{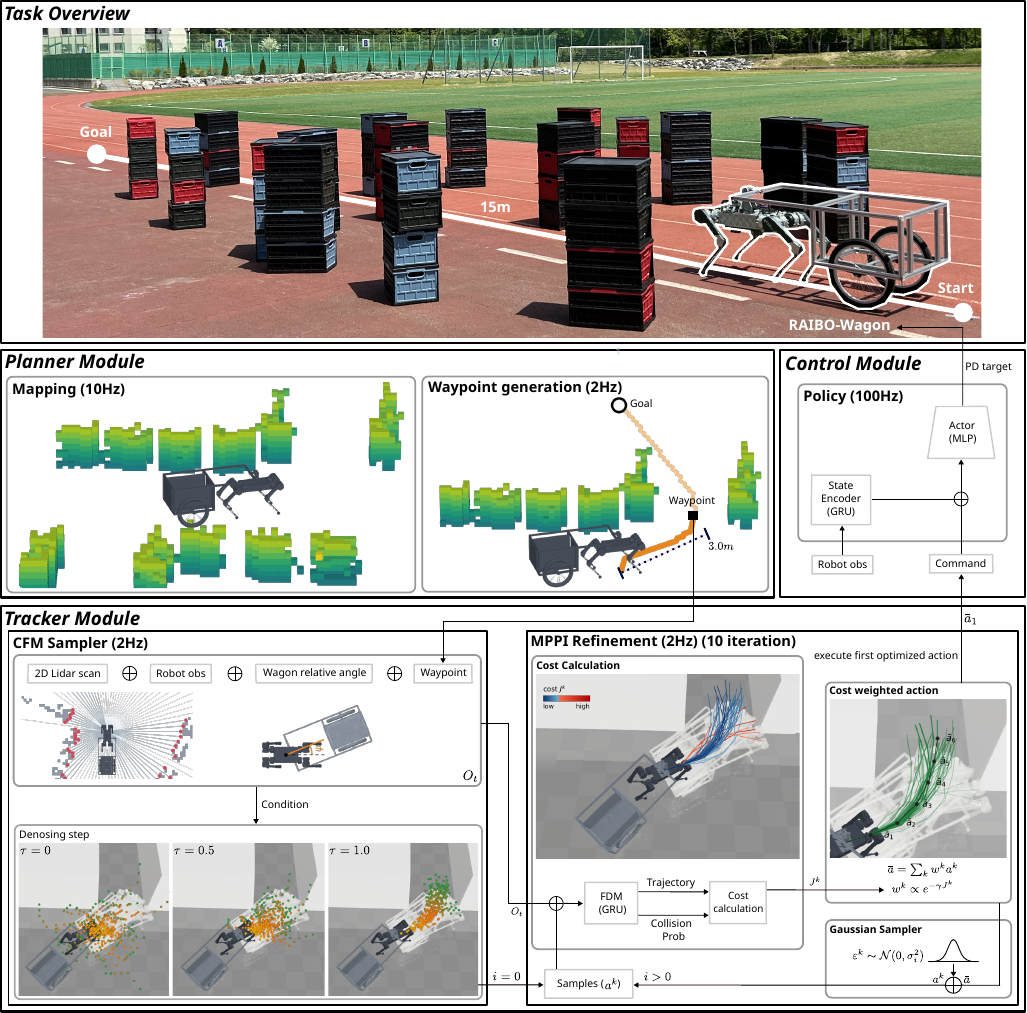}
    \caption{This figure illustrates the overall system executing real-time navigation with a quadruped robot towing a wagon. The robot first constructs a 3D occupancy map of the environment from onboard LiDAR. A lightweight A* planner then generates a global path, from which a waypoint is selected 3\,m ahead of the robot. The current navigation observation, comprising proprioceptive measurements, a LiDAR scan, and the relative yaw angle, is fed into a CFM model together with the waypoint to produce initial command samples for the MPPI optimizer. These warm-start samples are then iteratively refined by injecting decaying Gaussian noise and evaluating rollouts under a learned Forward Dynamics Model (FDM). After several iterations, the first command of the optimized sequence is issued to the low-level locomotion controller as an SE(2) velocity command.}
    \label{fig:new_method}
\end{figure*}

\section{Method}
We present a complete navigation system for the RAIBO-Wagon platform, spanning the full stack from low-level locomotion control to autonomous navigation. As illustrated in Fig.~\ref{fig:new_method}, the system maintains a real-time map of the environment and runs a global planner that supplies a waypoint to the local planner. Conditioned on this waypoint together with the navigation observation of the RAIBO-Wagon system, the CFM model produces a set of feasible candidate command sequences. These candidates are then refined over several iterations of Model Predictive Path Integral (MPPI) optimization, with each candidate evaluated by a learned Forward Dynamics Model (FDM), yielding the final optimized action sequence executed by the robot.

The following subsections build the system bottom-up, beginning with the locomotion controller (Sec.~\ref{sec:lowmethod}). We then present the expert navigation pipeline (Secs.~\ref{sec:expertplan} and~\ref{sec:experttrack}), which runs the same MPPI optimizer offline without real-time constraints; its solutions serve as the training data for the CFM model. Finally, we describe the local planner, in which the CFM model distilled from these solutions warm-starts real-time MPPI (Secs.~\ref{sec:learningcfm} and~\ref{sec:cfmw/mppi}).

\subsection{Low-Level Controller for Wagon Towing}
\label{sec:lowmethod}
To build a reliable wagon-towing navigation system, a locomotion controller specifically suited for wagon towing is required. The existing controller for RAIBO exhibits insufficient robustness when a wagon is attached, and its command-following precision degrades as the wagon mass varies. To address this, we modify the training details of the existing blind locomotion controller~\cite{lee2026quadruped} by introducing wagon-aware domain randomization into the reinforcement learning loop. We train the policy via Proximal Policy Optimization (PPO)~\cite{schulman2017proximal} in the RaiSim physics simulator~\cite{hwangbo2018per} with 200 parallel environments.

The policy consists of a GRU encoder and an MLP. The GRU encoder abstracts the robot state $o_t$, comprising body orientation, body angular velocity, joint positions, joint velocities, and the action history from the previous control step ($t - 0.01\,\mathrm{s}$). The MLP takes the resulting latent vector together with the commanded velocity and a standing-mode flag as input. Notably, no wagon-related state is provided to the policy: the controller remains entirely blind to the wagon and must infer its presence purely from proprioceptive feedback.

To improve command-following quality and robustness across varying wagon configurations, we apply wagon inertia randomization at each episode reset. The randomization follows a curriculum controlled by a factor $\alpha \in [0, 1]$ that progressively widens the range over training. The wagon mass is sampled uniformly from $[1.0,\ \max(1.5,\ 30.0\alpha)]\,\mathrm{kg}$ and the wheel mass from $[1.0,\ 1.0 + \alpha]\,\mathrm{kg}$. The center of mass is perturbed from its nominal position by offsets drawn from $[-0.1\alpha,\ 0.1\alpha]\,\mathrm{m}$ along each axis. The diagonal elements of the inertia tensor are scaled independently by factors sampled from $[1,\ 1 + \alpha]$, and the principal axes are rotated by angles drawn from $[-5\alpha,\ 5\alpha]^\circ$ about each axis.

\subsection{Expert Path Optimizer}
\label{sec:expertplan}
To collect training data for the CFM model, we construct an expert pipeline that generates feasible command sequences offline, free from real-time constraints. The expert pipeline consists of two components: a path optimizer that generates a collision-free path that accounts for the wagon, and a path tracker that follows this path without collision. In this section, we focus on the path optimizer, and in the following section we discuss the path tracker.

The most straightforward approach to path planning in a known map is the A* algorithm, which treats the robot as a circular footprint for collision checking. This assumption is reasonable for a standalone quadruped, whose geometry is well approximated by a circle, but it breaks down once a wagon is attached. Bounding the elongated quadruped-wagon system with a single circle greatly overestimates its footprint, yielding overly conservative paths in cluttered environments. Conversely, planning for the quadruped alone yields paths that may become infeasible with the wagon.

We therefore plan an initial A* path for the quadruped alone and refine it with a wagon-aware path optimizer. Rather than treating the quadruped-wagon system as a uniform circle, the optimizer explicitly accounts for the geometry and orientation of both the quadruped and the wagon, using a wagon-aware node cost described below. The refinement proceeds through repeated iterations of three steps: node cost evaluation, numerical gradient estimation, and path update.

The node cost consists of two terms: a learned risk cost related to the collision rate of the path and a smoothness cost related to the geometric smoothness of the path,
\begin{equation}
    \mathcal{C}_{\text{total}} = \mathcal{C}_{\text{risk}} + \mathcal{C}_{\text{smooth}}.
\end{equation}
The risk cost is computed by a learned risk predictor network $r_\theta$, which outputs a per-node risk score via a sigmoid activation,
\begin{equation}
    \mathcal{C}_{\text{risk}} = \sum_{i=1}^{N-1} \sigma\bigl(r_\theta(x_i)\bigr).
\end{equation}
The smoothness cost penalizes sharp directional changes along the path. At each node, the turning angle $\psi_i$ is defined as the difference between the forward and backward heading angles, wrapped to $[-\pi, \pi]$. The smoothness cost is then defined as
\begin{equation}
    \mathcal{C}_{\text{smooth}} = \sum_{i=1}^{N-1} \left(\frac{\psi_i}{\pi}\right)^2.
\end{equation}
The gradient of the total cost with respect to each node position is estimated using central differences. We adopt this finite-difference method because the heightmap input to the risk predictor is non-differentiable, precluding backpropagation through the network. The path is then updated in the direction of the negative gradient, and this procedure is repeated for 25 iterations per planning call.

The learned risk predictor network is trained on data collected by rolling out RAIBO-Wagon in the RaiSim physics simulator. The risk predictor consists of a CNN block and an MLP. The CNN block encodes a heightmap patch into a latent vector. The MLP takes this latent vector together with the node state, comprising the $x$ and $y$ coordinates and the orientations of the robot and the wagon, and predicts the collision score. The risk predictor is trained using binary cross-entropy loss,
\begin{equation}
    \mathcal{L}_{\text{risk}} = -\sum_{i} 
    \bigl[ y_i \log \hat{y}_i 
    + (1 - y_i)\log(1 - \hat{y}_i) \bigr],
\end{equation}
where $y_i \in \{0, 1\}$ is the ground-truth collision label.

\subsection{Expert Path Tracker}
\label{sec:experttrack}
When a wagon is attached, a simple kinematic path tracker deviates further from the planned collision-free path, as the coupled quadruped-wagon dynamics make exact path following infeasible. In cluttered environments in particular, this deviation can lead to collisions. A dedicated tracker is therefore needed that both follows the path accurately and keeps the quadruped-wagon system collision-free. Meeting both requirements calls for explicit lookahead over future states, which neither a reactive feedback controller nor an RL tracker provides. We therefore instantiate the MPPI formulation of Sec.~\ref{sec:preliminary_mppi} with a learned Forward Dynamics Model (FDM)~\cite{kim2022learning, roth2025learned} as the rollout model for our expert path tracker. The tracker optimizes over a horizon of $T=6$ steps with $K=1024$ candidates and 25 iterations.

\textbf{Forward Dynamics Model (FDM).} 
One of the most important parts of MPPI is the dynamics model used to roll out candidate command sequences during optimization. Several choices are available, such as a simple kinematic model or a full dynamics model from a rigid-body simulator. A simple kinematic model, however, cannot capture the quadruped-wagon system accurately enough for safe navigation, while a full dynamics model is computationally prohibitive under the real-time budget. Following~\cite{kim2022learning, roth2025learned}, we therefore train a neural network FDM that predicts the future trajectory and a per-step collision logit of the coupled system given the current observation and a candidate command sequence. We further evaluate the effectiveness of the FDM in terms of computational time and navigation performance in the Appendix. 

The model consists of four components: a state encoder, a command encoder, a GRU, and two decoders. Given the current navigation observation $o \in \mathbb{R}^{133}$---including LiDAR observations, proprioceptive observations, and the robot-wagon relative angle---the state encoder $f_{\text{state}}$ produces a latent vector that initializes the GRU hidden state,
\begin{equation}
    h_0 = f_{\text{state}}(o), \quad h_0 \in \mathbb{R}^{128}.
\end{equation}
At each timestep $t$, the command encoder $f_{\text{cmd}}$ encodes the command $a_t \in \mathbb{R}^{3}$,
\begin{equation}
    z_t = f_{\text{cmd}}(a_t), \quad z_t \in \mathbb{R}^{128},
\end{equation}
which is then passed through the GRU to produce a hidden state $h_t$. The trajectory decoder $f_{\text{traj}}$ and collision decoder $f_{\text{coll}}$ map $h_t$ to the predicted pose increment and collision logit, respectively,
\begin{equation}
    \Delta \hat{p}_t = f_{\text{traj}}(h_t) \in \mathbb{R}^{4}, \qquad \hat{d}_t = f_{\text{coll}}(h_t) \in \mathbb{R}.
\end{equation}
Here, $\Delta \hat{p}_t$ comprises the per-step position and heading increments of the robot together with the increment of the robot-wagon relative angle, expressed in the robot frame at the current planning instant. The poses entering the reward, $\hat{p}_t^{r}$, $\hat{\theta}_t^{r}$, $\hat{p}_t^{w}$, and $\hat{\theta}_t^{w}$, are recovered by accumulating these increments along the horizon; the wagon pose then follows from the robot pose and the relative angle.

The FDM is trained with a combined trajectory regression loss and a class-reweighted binary cross-entropy collision loss,
\begin{equation}
    \mathcal{L}_{\text{FDM}} = \mathcal{L}_{\text{traj}} + 0.1 \cdot \mathcal{L}_{\text{coll}},
\end{equation}
where $\mathcal{L}_{\text{traj}}$ is the MSE between the predicted and ground-truth pose increments of the robot and the robot-wagon relative angle, and $\mathcal{L}_{\text{coll}}$ is a weighted BCE loss over episode termination events.

To train the FDM, we collect data via iterative online rollouts in the RaiSim physics simulator, with the environment and platform parameters randomized at each episode reset. Table~\ref{tab:fdm_randomization} summarizes the randomization ranges. We found that this randomization not only improves sim-to-real transfer quality but also encourages conservative path tracking. In total, we collect approximately 500{,}000 episodes.

\begin{table}[h]
\centering
\caption{FDM Training Randomization Ranges.}
\label{tab:fdm_randomization}
\begin{tabular}{lcc}
\hline
Parameter & Range & Unit \\
\hline
Wagon mass         & $[1.0,\ 30.0]$       & kg \\
Wheel mass         & $[1.0,\ 2.0]$        & kg \\
COM offset         & $[-0.1,\ 0.1]$       & m  \\
Inertia scale      & $[1.0,\ 2.0]$        & --  \\
Wagon size         & $[0.9,\ 1.1]\times$  & -- \\
Terrain roughness  & $[0.0,\ 0.05]$       & m  \\
Terrain slipperiness & $[0.3,\ 1.0]$      & -- \\
\hline
\end{tabular}
\end{table}

\textbf{Reward Evaluation.} Each candidate sequence is rolled out through the FDM, which predicts the future trajectories of both the robot and the wagon, as well as a per-step collision logit. The tracking losses are computed as weighted sums over the horizon,
\begin{align}
    \mathcal{L}_{\text{pos}}  &= \sum_{t=1}^{T} \lambda^t \left( \frac{\| p_t^{r,\text{ref}} - \hat{p}_t^{r} \|_2}{\tau_r} + \frac{\| p_t^{w,\text{ref}} - \hat{p}_t^{w} \|_2}{\tau_w} \right), \\
    \mathcal{L}_{\text{ori}}  &= \sum_{t=1}^{T} \lambda^t \left( \frac{| \theta_t^{r,\text{ref}} - \hat{\theta}_t^{r} |}{\tau_{\theta_r}} + \frac{| \theta_t^{w,\text{ref}} - \hat{\theta}_t^{w} |}{\tau_{\theta_w}} \right), \\
    \mathcal{L}_{\text{coll}} &= \sum_{t=1}^{T} c_t \cdot \log(1 + e^{\hat{d}_t}).
\end{align}
The reference terms $p_t^{r,\text{ref}}$, $\theta_t^{r,\text{ref}}$, $p_t^{w,\text{ref}}$, and $\theta_t^{w,\text{ref}}$ are obtained by resampling the optimized path at $0.5\,\mathrm{m}$ intervals, matching the per-step displacement of the tracker. Since each node of this path carries the full quadruped-wagon configuration $(x_i,\, y_i,\, \theta_i^{r},\, \theta_i^{w})$, the wagon references come directly from the wagon-aware optimizer (Sec.~\ref{sec:expertplan}) rather than being inferred from a robot-only path. Here, $\lambda$ is a per-step discount factor, $\tau$ denotes temperature scaling for each component, and $c_t \in \{1.0, 1.0, 0.9, 0.9, 0.8, 0.8\}$ is a time-decaying collision weight that down-weights distant collision predictions. The total reward is then defined as

\begin{equation}
    R = e^{-\mathcal{L}_{\text{pos}}} + 0.5\, e^{-\mathcal{L}_{\text{ori}}} + e^{-\mathcal{L}_{\text{coll}} / 15}.
\end{equation}

\subsection{Learning a Conditional Flow Matching Model}
\label{sec:learningcfm}

\begin{algorithm}[t]
\caption{Conditional Flow Matching Training with CFG}
\label{alg:cfm_training}
\begin{algorithmic}[1]
\REQUIRE Environment $\mathcal{E}$, A* planner $\pi_{\text{A*}}$, expert path optimizer $\pi_{\text{opti}}$, expert path tracker $\pi_{\text{track}}$, model $v_\phi$, drop probability $p_{\text{drop}}$, learning rate $\eta$
\STATE Initialize dataset $\mathcal{D} \leftarrow \emptyset$
\STATE \textbf{// Data collection}
\FOR{each episode}
    \STATE Sample goal $g$ at a distance of $15$--$20\,\mathrm{m}$ from start
    \STATE Query $\pi_{\text{opti}}$ to plan a wagon-aware optimized path $\mathcal{P}$
    \WHILE{goal $g$ not reached}
        \STATE Query $\pi_{\text{A*}}$ to obtain raw path and waypoint $g_{\text{way}}$  
        \STATE Execute $\pi_{\text{track}}$ to follow $\mathcal{P}$ and obtain expert command sequence $x_{\text{exp}}$
        \STATE Observe navigation observation $o$
        \STATE $\mathcal{D} \leftarrow \mathcal{D} \cup \{(o,\, g_{\text{way}},\, x_{\text{exp}})\}$ \hfill $\triangleright$ every $1.5\,\mathrm{s}$
    \ENDWHILE
\ENDFOR
\STATE \textbf{// Learning Conditional Flow Matching model}
\REPEAT
    \STATE Sample mini-batch $\{(o,\, g_{\text{way}},\, x_{\text{exp}})\} \sim \mathcal{D}$
    \STATE Sample prior $\epsilon \sim \mathcal{N}(\mathbf{0}, \mathbf{I})$
    \STATE Sample flow time $t = \operatorname{sigmoid}(z),\ z \sim \mathcal{N}(0, 1)$
    \STATE Interpolate $x_t \leftarrow (1 - t)\, \epsilon + t\, x_{\text{exp}}$
    \STATE Sample CFG mask $d \sim \text{Bernoulli}(p_{\text{drop}})$
    \STATE Apply dropout $c \leftarrow (1 - d) \cdot [o;\, g_{\text{way}}]$
    \STATE Compute loss $\mathcal{L} \leftarrow \| v_\phi(x_t, c, t) - (x_{\text{exp}} - \epsilon) \|^2$
    \STATE Update $\phi \leftarrow \phi - \eta\, \nabla_\phi \mathcal{L}$
\UNTIL{converged}
\end{algorithmic}
\end{algorithm}

The expert pipeline (Secs.~\ref{sec:expertplan} and~\ref{sec:experttrack}) produces dynamically feasible command sequences for the quadruped towing a wagon, but only offline. Its optimization requires a computational budget far beyond what real-time operation allows. We therefore distill its solutions into a generative model, which has proven effective at capturing multimodal trajectory distributions~\cite{chi2025diffusion}. Specifically, we adopt Conditional Flow Matching (CFM)~\cite{lipman2022flow} with Classifier-Free Guidance (CFG)~\cite{ho2022classifier}.

\textbf{Model and Objective.}
The CFM model $v_\phi$ consists of a condition encoder and a main network.
The condition encoder is a two-layer MLP that projects the conditioning variable $c$ into a latent vector $z_c \in \mathbb{R}^{128}$.
The flow time $t \in [0,1]$ is encoded via a sinusoidal positional embedding into $z_t \in \mathbb{R}^{64}$.
The main network is a four-layer MLP with hidden dimensions $[512, 512, 256, 256]$ and LeakyReLU activations, taking as input the concatenation $[x_t;\, z_c;\, z_t]$ and outputting the predicted vector field.
At each planning step, the conditioning variable is defined as $c = [o;\, g_{\text{way}}]$. Here, $o \in \mathbb{R}^{133}$ is the current navigation observation, including LiDAR observations, the robot-wagon relative angle measured by an encoder, and the robot's proprioceptive state, and $g_{\text{way}} \in \mathbb{R}^{2}$ is a waypoint located $3\,\mathrm{m}$ ahead along the planned A* path. Rather than conditioning on the full path, we condition on a local waypoint, which encourages the model to focus on immediate feasibility. Global consistency is maintained by the A* planner, which continuously updates $g_{\text{way}}$ as the robot progresses. The model is trained following the CFG procedure introduced in Sec.~\ref{sec:cfg_training}, with $p_{\text{drop}} = 0.1$. We sample the flow time as $t = \operatorname{sigmoid}(z)$, $z \sim \mathcal{N}(0,1)$, rather than uniformly, concentrating training on intermediate $t$.

\textbf{Data Collection.}
We collect the dataset for the CFM model using the expert path optimizer and path tracker introduced in Secs.~\ref{sec:expertplan} and~\ref{sec:experttrack}. In every episode, the goal position is sampled in a random direction at a distance of 15--20\,m from the start. The expert path optimizer plans a wagon-aware path to the goal, and the path tracker follows it. We roll out this pipeline in the RaiSim physics simulator, randomizing the environment at each episode in the same manner as for the FDM (Table~\ref{tab:fdm_randomization}). Every 1.5\,s during a rollout, we record a training sample consisting of the navigation observation $o$, the waypoint $g_{\text{way}}$, and the optimized expert command sequence $x_{\text{exp}}$ of length 6 (3\,s). Note that $g_{\text{way}}$ is extracted from a raw A* path rather than from the expert-optimized path, matching the input available at deployment time, where the expert optimizer cannot run. In total, we roll out 25{,}000 episodes and collect approximately 250{,}000 samples; the full procedure is summarized in Algorithm~\ref{alg:cfm_training}.

\subsection{Conditional Flow Matching Model with MPPI}
\label{sec:cfmw/mppi}
\begin{algorithm}[t]
    \caption{CFM-Initialized MPPI}
    \label{alg:flow_fdm}
    \begin{algorithmic}[1]
        \REQUIRE Current observation $o_t$, waypoint $g_{\text{way}}$,
                 FDM $f_\theta$, CFM model $v_\phi$, guidance weight $w$,
                 ODE steps $N$, iterations $I$, samples $K$
        \STATE \textbf{// CFM Inference with CFG}
        \STATE Set condition $c \leftarrow [o_t;\, g_{\text{way}}]$
        \STATE $\tilde{v}(x, t) \leftarrow w\,v_\phi(x, c, t) + (1-w)\,v_\phi(x, \emptyset, t)$
        \FOR{$k = 1$ \TO $K$}
            \STATE $x_0^{k} \sim \mathcal{N}(\mathbf{0}, \mathbf{I})$
            \FOR{$n = 0$ \TO $N-1$}
                \STATE $x_{n+1}^{k} \leftarrow x_n^{k} + \tilde{v}(x_n^{k}, n/N) / N$
            \ENDFOR
        \ENDFOR
        \STATE \textbf{// Iterative MPPI with FDM}
        \FOR{$i = 1$ \TO $I$}
            \STATE {// from CFM model}
            \IF{$i = 1$}
                \STATE Set $\mathbf{a}^{k} \leftarrow x_N^{k}$ for all $k$
            \ENDIF
            \STATE {// from previously optimized actions}
            \IF{$i > 1$}
                \STATE $\mathbf{a}^{k} \leftarrow \bar{\mathbf{a}} + \boldsymbol{\epsilon}^{k},\quad \boldsymbol{\epsilon}^{k} \sim \mathcal{N}(0,\,\sigma_i^2\,\Sigma_\beta)$
            \ENDIF
            \FOR{$k = 1$ \TO $K$}
                \STATE $\hat{\tau}^{k}, \hat{\mathbf{d}}^{k} \leftarrow f_{\theta}(o_t, \mathbf{a}^{k})$ \hfill $\triangleright$ rollout
                \STATE $R^{k} \leftarrow R(\hat{\tau}^{k}, \hat{\mathbf{d}}^{k}, g_{\text{way}})$ \hfill $\triangleright$ Eq.~(\ref{eq:online_reward})
            \ENDFOR
            \STATE $\bar{\mathbf{a}} \leftarrow
                \dfrac{\sum_{k=1}^{K} \exp\!\bigl(\gamma (R^{k} - R_{\max})\bigr)\, \mathbf{a}^{k}}
                      {\sum_{k=1}^{K} \exp\!\bigl(\gamma (R^{k} - R_{\max})\bigr)}$ \hfill $\triangleright$ optimizing
        \ENDFOR
        \STATE Apply first action $\bar{a}_1$ to the robot
    \end{algorithmic}
\end{algorithm}

The learned CFM model meets the real-time budget, generating dynamically plausible command sequences in only ten integration steps. However, we observe that using the CFM model alone substantially degrades navigation performance relative to the offline expert: as a distilled approximation, it cannot fully reproduce the expert's optimization. We therefore refine the CFM samples online with a lightweight MPPI stage, which not only closes this gap but also surpasses the offline expert. We attribute this to the training data: the CFM model is distilled only from expert solutions that converged successfully, so its prior concentrates on the successful subset of expert behavior.

Our local planner operates in two phases. The CFM model first generates 256 candidate sequences, and MPPI then iteratively refines them through Gaussian exploration, evaluating each candidate with the learned FDM. The full procedure is summarized in Algorithm~\ref{alg:flow_fdm}.

\textbf{Phase 1: CFM Warm-Start.}
In the first phase of optimization, our learned CFM model generates 256 independent command samples via CFG-guided ODE integration (lines~3--9). All 256 samples are used directly as the first-iteration command set for MPPI. CFG biases each sample toward the waypoint direction while preserving diversity across samples, yielding a warm-started yet exploratory initialization. The guidance weight $w$ controls this trade-off: too large a value collapses diversity, while too small a value weakens the directional bias. Empirically, we find $w = 3$ to strike a favorable balance, and use this value throughout our experiments.

\textbf{Phase 2: Iterative MPPI with FDM.}
Starting from the CFM-initialized action set, we run 10 iterations of MPPI using the learned FDM $f_\theta$ to evaluate each candidate. From the second iteration onward, temporally correlated Gaussian noise with decaying standard deviation $\sigma_i$ is added around the current mean $\bar{\mathbf{a}}$ (line~18). The FDM predicts the future trajectories of both the robot and the wagon, $\hat{\tau}^{k}$, together with per-step collision logits $\hat{\mathbf{d}}^{k}$, in a single forward pass. Unlike the expert tracker of Sec.~\ref{sec:experttrack}, which tracks a wagon-aware reference path, the local planner has access only to the raw A* path. The path optimizer is excluded because it exceeds the real-time budget. We therefore score candidates against the same local waypoint $g_{\text{way}}$ that conditions the CFM model, combining waypoint proximity with the predicted collision cost,
\begin{equation}
    R = e^{-\mathcal{L}_{\text{way}}} + e^{-\mathcal{L}_{\text{coll}}/15},
    \label{eq:online_reward}
\end{equation}
where $\mathcal{L}_{\text{way}} = \| g_{\text{way}} - \hat{p}_T^{r} \|_2 / \tau_g$ and $\mathcal{L}_{\text{coll}}$ is defined as in Sec.~\ref{sec:experttrack}. Note that the wagon enters this reward through $\mathcal{L}_{\text{coll}}$ alone: because the FDM predicts the wagon trajectory jointly with the robot's, wagon-obstacle contact is penalized without requiring a wagon reference pose. A softmax-weighted update then refines $\bar{\mathbf{a}}$ toward high-reward regions. This schedule follows a coarse-to-fine strategy: early iterations with large $\sigma_i$ broadly explore the feasible region established by the warm-start, while later iterations with small $\sigma_i$ precisely converge to the final action sequence.

\textbf{Complementarity of the Two Phases.}
The two phases are complementary by design. The CFM model restricts the initial sample set to a high-reward region that previous-action initialization cannot reach. Iterative MPPI, in turn, explores around that region through Gaussian perturbations and reaches solutions outside the CFM model's training support, addressing the CFM model's distributional ceiling.

\textbf{Addressing Multimodality.}
However, a naive instantiation of this scheme fails because the CFM model is inherently multimodal: given the same observation, it may produce qualitatively distinct avoidance trajectories corresponding to different modes. Averaging such samples as the MPPI initialization collapses the distribution to a meaningless intermediate, leading to poor optimization. Prior work~\cite{mizuta2025unified} addresses this by selecting a few high-reward samples and running a separate MPPI branch on each. However, this multiplies the computational cost by the number of branches and does not scale to real-time operation. We instead condition the CFM model on a waypoint extracted from the A* path, restricting the sample distribution to a single navigable corridor and effectively collapsing it to a unimodal regime. This mechanism also enables reliable replanning: when the original path is blocked, the updated A* waypoint redirects the sample distribution toward a new corridor without any additional logic.

\section{Implementation Details}

For the real-world experiments, we built a custom wagon designed to be towed by a quadruped. Designing the wagon for the quadruped, rather than adapting an off-the-shelf one, offers two key advantages. First, we embed an angle encoder directly into the yaw joint of the connector, providing a reliable measurement of the relative yaw angle between the quadruped and the wagon. This avoids reliance on vision or other exteroceptive-sensor-based estimation, which inevitably introduces error. Second, sizing the wagon to match RAIBO's form factor minimizes kinematic interference during locomotion and prevents collisions when the robot stands up or sits down.

\begin{figure}[!t]
    \centering
    \includegraphics[width=1.0\linewidth]{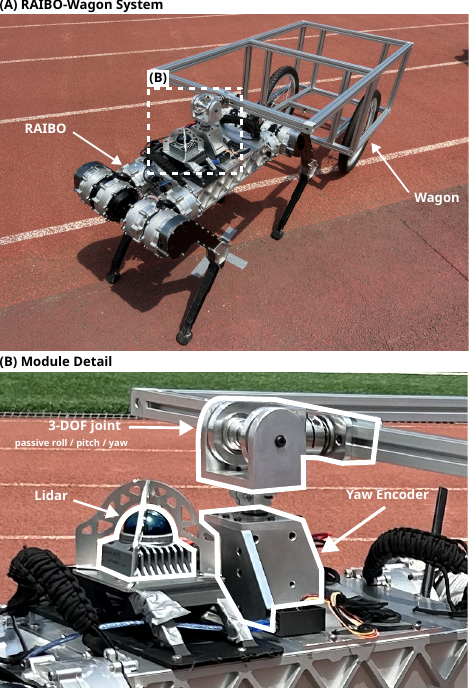}
    \caption{This figure illustrates the hardware design of RAIBO-Wagon. (A) shows the fully integrated platform with the robot and the wagon. (B) highlights the key modules mounted on the robot: the Livox MID-360 LiDAR and the wagon connector. The wagon connector consists of a 3-DOF passive joint that accommodates all possible 3D relative angles between the robot and the wagon, with a dedicated angular encoder measuring the yaw angle.}
    \label{fig:hardware}
\end{figure}

The overall system is shown in Fig.~\ref{fig:hardware}(A). The wagon is fabricated from extruded aluminum profiles. Fig.~\ref{fig:hardware}(B) details the remaining modules. A Livox MID-360 LiDAR is mounted at the front of the robot and plays several roles: localization, mapping, and scan input for the navigation framework. The wagon and robot are coupled through a connector with three passive joints,
permitting rotation of the wagon relative to the robot about all three axes. The joint governing the relative yaw is equipped with an angle encoder, enabling direct measurement of the relative yaw angle.

Our navigation system is organized into three modules---mapping and localization, command generation, and low-level control---detailed below.

\begin{figure*}[!t]
    \centering
    \includegraphics[width=0.95\linewidth]{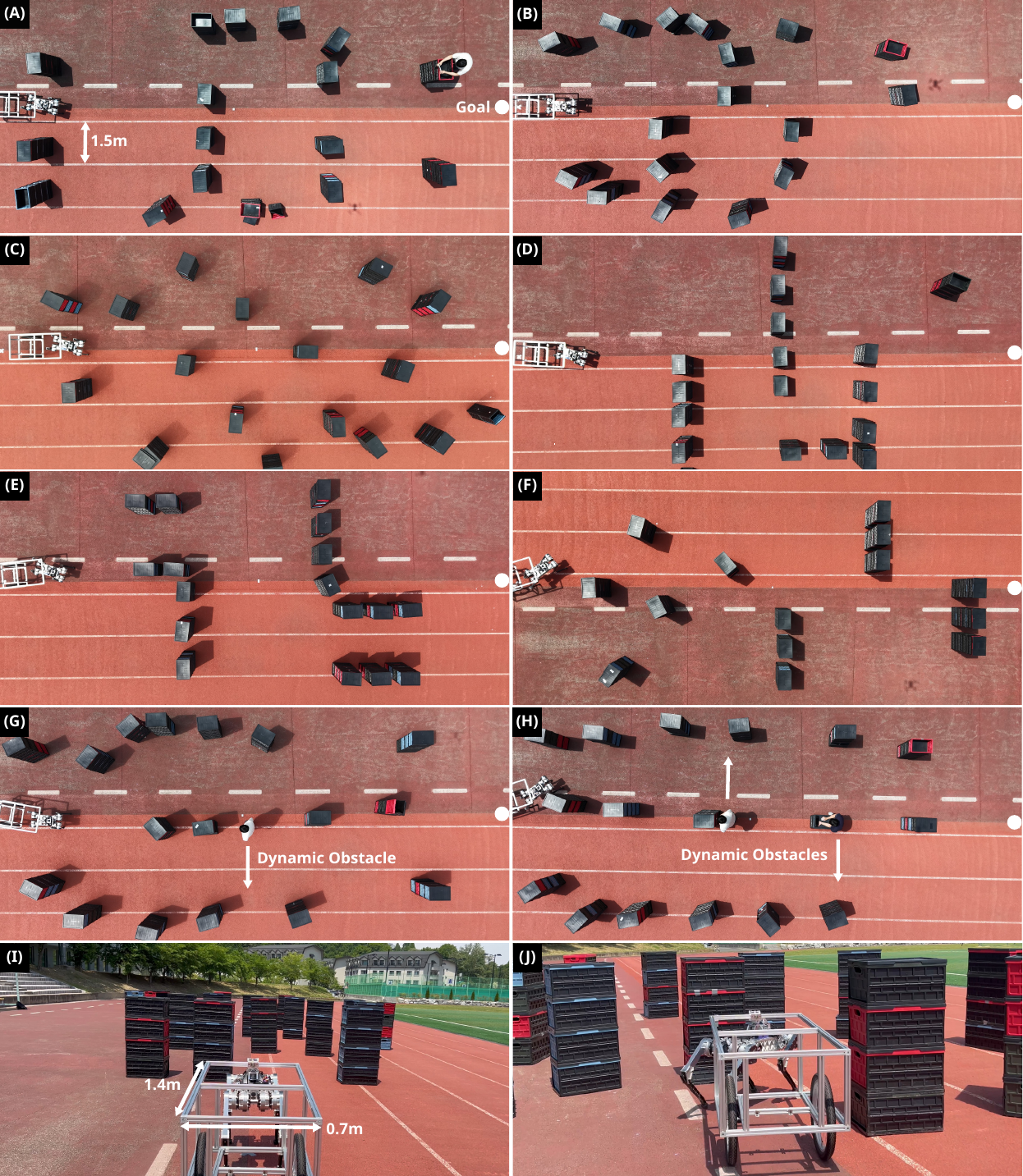}
    \caption{This figure shows an overview of the real-world experiment settings. Each trial consists of a 15\,m real-time navigation task. (A) through (F) are static obstacle settings, in which the obstacles remain fixed throughout the trial. (G) and (H) are dynamic obstacle settings in which humans act as moving obstacles; the white arrows indicate their walking directions. (I) and (J) are snapshots from our real-world experiments. In (A), one grid step of the obstacle layout corresponds to 1.5\,m, and (I) shows the wagon, which measures 0.7\,m $\times$ 1.4\,m.}
    \label{fig:terrain}
\end{figure*}

\noindent\textbf{Mapping and Localization.}
LiDAR point clouds are voxelized using 5\,cm voxels and fused into a 3D occupancy map at 10\,cm resolution using OctoMap~\cite{hornung2013octomap} at 10\,Hz, with robot localization supplied by FAST-LIO2~\cite{xu2022fast} running at the same rate. However, mapping in our setting requires one additional consideration. The wagon would otherwise be registered as an obstacle in the map, corrupting path planning; we prevent this by masking the region it occupies, using the relative yaw angle measured by the wagon connector encoder.

\noindent\textbf{Command Generation.}
For path planning, we project a 2.5D grid map from the accumulated OctoMap. An A* global planner then generates a reference path over this map at 2\,Hz and passes waypoints to the local planner. Running at the same rate, the local planner produces a sequence of 6 commands over a 3\,s horizon each cycle, corresponding to a target navigation speed of 1\,m/s (0.5\,m per step). Internally, it warm-starts from the CFM model with CFG ($\approx$0.15\,s) and refines the resulting samples with 10 MPPI iterations under the learned forward dynamics model ($\approx$0.15\,s), for a total of 0.3\,s per planning cycle.

\begin{figure*}[!t]
    \centering
    \includegraphics[width=0.95\linewidth]{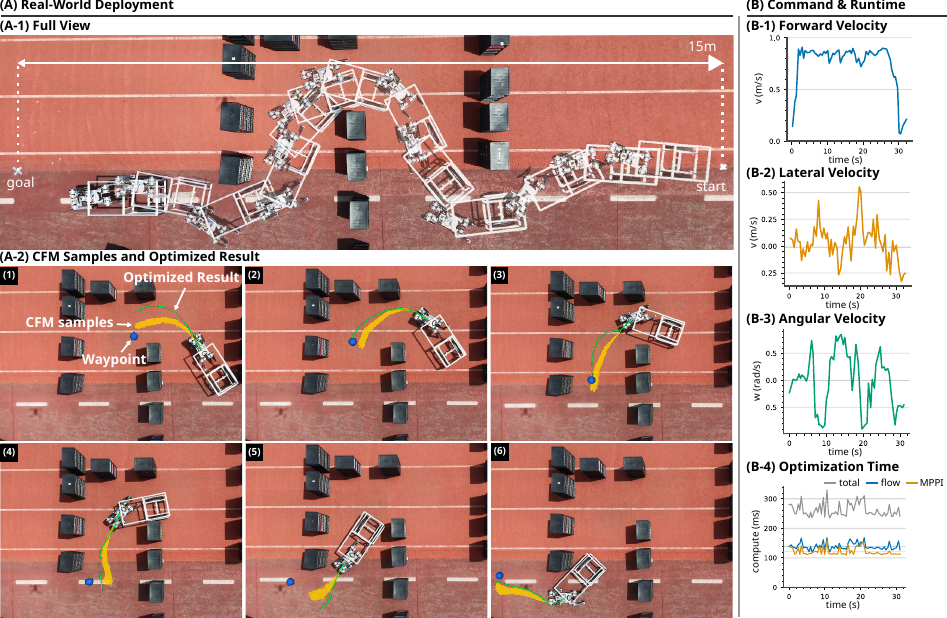}
    \caption{(A) shows real-world deployment snapshots captured by a drone. (A-1) shows the overall navigation trajectory on Terrain D. (A-2) shows six examples of the initial CFM model samples and the final optimized result, labeled (1) through (6), on the same terrain. The blue dot is the waypoint that is provided to both the CFM model and MPPI. The yellow lines are the 256 samples from the CFM model, and the green line is the final command sequence optimized by MPPI. (B) shows the commands output by the local planner and its runtime. (B-1) through (B-3) show the forward, lateral, and angular velocities of the optimized command, respectively. (B-4) shows the onboard runtime of each planning cycle: the CFM sampling time, the MPPI refinement time, and their total.}
    \label{fig:result1}
\end{figure*}

\noindent\textbf{Control.}
The RL-based low-level locomotion controller operates at 100\,Hz, consuming the first command from the optimized command sequence output by the command generation module.

The entire navigation system runs onboard RAIBO using a NUC 15 Pro, with no offboard processing.

\section{Results}

We evaluate our navigation system in both real-world and simulation settings. In the real-world experiments, we analyze whether our system can execute safe autonomous navigation across diverse terrain configurations. In simulation, we conduct an ablation study and compare against other methods. We first present the real-world experiment results, followed by the ablation study.

\subsection{Real-World Evaluation}

\begin{figure*}[t]
    \centering
    \includegraphics[width=0.95\linewidth]{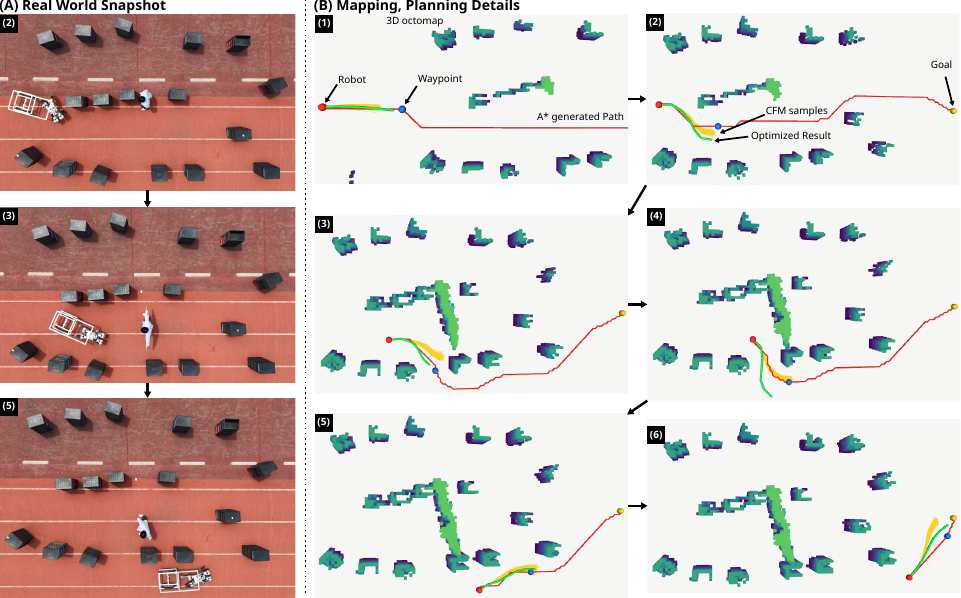}
    \caption{This figure shows the overall navigation process in a dynamic obstacle setting. (A) shows drone snapshots during autonomous navigation, taken at scenes (2), (3), and (5) of (B). (B) shows the mapping and planning details in the same setting: the boxes are voxels of the 3D OctoMap, colored by height from purple at ground level to green at the top, the red line is the path generated by the A* planner from that map, the blue dot is the waypoint sampled from the path, the red dot is the current robot position, the yellow dot is the goal position, and the yellow and green lines are the CFM samples and the optimized result, respectively.}
    \label{fig:replanning}
\end{figure*}

\subsubsection{Experimental Setup}

We evaluate our system on a 15\,m course without any prior map. The evaluation covers eight obstacle configurations, six static (Terrains A--F) and two dynamic (Terrains G and H), with five trials per configuration. Fig.~\ref{fig:terrain} illustrates the experimental configurations. In the dynamic settings, humans act as moving obstacles, and the white arrows indicate their walking directions. Specifically, Terrain G contains a single human who blocks one of the two available routes, whereas Terrain H contains two humans who block both routes simultaneously. These settings evaluate the system's responsiveness to environmental changes that occur during navigation.

\subsubsection{Success Rate and Failure Analysis}

For each terrain, we conduct five trials to evaluate the success rate and analyze failure modes. As shown in Table~\ref{tab:failure_analysis}, our system achieves an overall success rate of 85.0\% (34/40) across the eight configurations.

The six failures split evenly between two modules: three arise in mapping, which builds the environment representation, and three in the local planner, which follows the waypoints generated by the global planner.

The mapping failures share a common cause. To keep the wagon from being registered as an obstacle, we mask the rear 180\textdegree\ of the LiDAR's field of view. During sharp turns, however, the wagon swings into the robot's lateral field of view. The mask then also removes regions that contain real obstacles, degrading the occupancy map precisely in the most safety-critical situations. Consistent with this mechanism, all three mapping failures occurred in the terrains with the sharpest turns (C, D, and H). In every case, the obstacle was missed entirely, and the robot towed the wagon straight into it, a clear collision rather than a near-miss.

The remaining three failures stem from the local planner. Although several factors can contribute to planning error, we identify the dominant cause as a sim-to-real gap in the 2D LiDAR input supplied to the CFM model and the FDM. In training, these models observe a complete, instantaneous 2D scan, whereas at deployment they do not. We feed raw MID-360 scans directly to the local planner instead of reconstructing the 2D scan from the occupancy map, since reconstruction would compound the error introduced by the wagon masking. The direct MID-360 input, however, has a limitation of its own. Its non-repetitive scan pattern covers the full 360\textdegree\ field of view only over a finite integration window, so a single scan can contain stale or missing returns. In confined spaces, these missing detections lead the local planner to act on an outdated obstacle configuration. Unlike the mapping failures, these cases were marginal. The wagon only grazed the obstacle rather than driving into it, indicating that the local planner had nearly resolved the avoidance. Both failure modes thus trace back to the wagon attachment. The mapping failures stem directly from the masking it requires, and the planning failures stem indirectly from the resulting reliance on a raw sensor stream whose staleness the models never encountered in training.

\begin{table}[t]
\renewcommand{\arraystretch}{1.25}
\caption{Success rate and failure case analysis.}
\label{tab:failure_analysis}
\centering
\begin{tabular}{l|cccccccc}
\hline\hline
 & \textbf{A} & \textbf{B} & \textbf{C} & \textbf{D} & \textbf{E} & \textbf{F} & \textbf{G} & \textbf{H} \\
\hline
Success            & 5/5  & 4/5  & 4/5  & 4/5  & 4/5  & 5/5  & 5/5 & 3/5 \\
Fail (Mapping)  & --   & 0/1  & 1/1  & 1/1  & 0/1  & --  & -- & 1/2 \\
Fail (Planning) & --   & 1/1  & 0/1  & 0/1  & 1/1  & --  & -- & 1/2 \\
\hline\hline
\end{tabular}
\end{table}

\subsubsection{Overcoming Confined Obstacles}

The test courses are highly confined: as shown in Fig.~\ref{fig:result1}(A-1) for Terrain D, RAIBO-Wagon navigates through obstacles with only a narrow margin. Even in such confined spaces, our system achieves an 85\% success rate across all eight configurations. To illustrate how this is achieved, we examine the behavior of our framework on Terrain D in detail.

Fig.~\ref{fig:result1}(A-1) shows the full navigation path, and Fig.~\ref{fig:result1}(A-2) overlays the initial CFM samples with the final command trajectory refined by MPPI. We observe two patterns. First, both the CFM samples and the final trajectory deviate from the A* waypoints whenever the direct path is unsafe, reflecting the collision awareness of the framework. Second, the CFM model alone is insufficient in tight regions, as in scenes~(1) and~(3) of Fig.~\ref{fig:result1}(A-2). If RAIBO-Wagon followed the CFM samples there, the wagon would collide with the obstacle. MPPI refinement corrects these cases by explicitly accounting for the wagon through the learned FDM. This enables the framework to pass through confined obstacle settings with the bulky wagon attached.

\subsubsection{Real-Time Navigation}

Unlike Terrains A--F, which contain only static obstacles, Terrains G and H introduce dynamic obstacles in the form of moving humans.

Fig.~\ref{fig:replanning}(A) illustrates the overall behavior. The robot initially selects one of two available paths and begins navigation. When a human steps in to block this
path, the robot detects the new obstacle through online mapping and replans to the alternative route, completing the navigation without collision. Fig.~\ref{fig:replanning}(B)
provides a detailed view of the mapping and planning evolution. In scene~(1), prior to detecting the obstacle, the robot proceeds along the original path. From scene~(2), as the obstacle enters the sensing range, the occupancy map is updated and the planned path is slightly adjusted. In scene~(3), once the path is fully blocked, the A* planner replans and places a new waypoint along the unobstructed route, which the local planner then follows. After replanning, the robot successfully reaches the goal, as can be seen in scenes (4) through (6).

This behavior confirms that the system can navigate without a prior map by combining online mapping, global replanning, and real-time command optimization.

\subsubsection{Additional Mass Robustness}

\begin{table}[t]
\renewcommand{\arraystretch}{1.25}
\caption{Success rate with regard to the additional wagon mass.}
\label{tab:wagonrandom}
\centering
\begin{tabular}{l|ccc}
\hline\hline
 & \textbf{+5\,kg} & \textbf{+10\,kg} & \textbf{+15\,kg} \\
\hline
Success            & 5/5  & 5/5  & 5/5 \\
\hline\hline
\end{tabular}
\end{table}

We also show that our system is robust to varying wagon payloads. We evaluate this robustness by mounting additional payloads of 5, 10, and 15\,kg atop the wagon in Terrain F and repeating the trials. As shown in Table~\ref{tab:wagonrandom}, the success rate remains 100\% across all payload conditions. Note that the base wagon itself weighs approximately 25\,kg, so the robot is pulling up to 40\,kg in the heaviest condition.

We do not explicitly provide the wagon's mass or center of mass to either the
locomotion controller or the local planner. Instead, our system implicitly accommodates the varying wagon mass through proprioceptive information. This adaptation stems from two complementary sources: (i) the wagon-specific locomotion controller maintains a stable gait under varying inertial loads, as detailed in the Appendix, and (ii) both the CFM model and the FDM are trained with randomized wagon mass in simulation, yielding robustness under payload variation.

\begin{figure}[!t]
    \centering
    \includegraphics[width=1.0\linewidth]{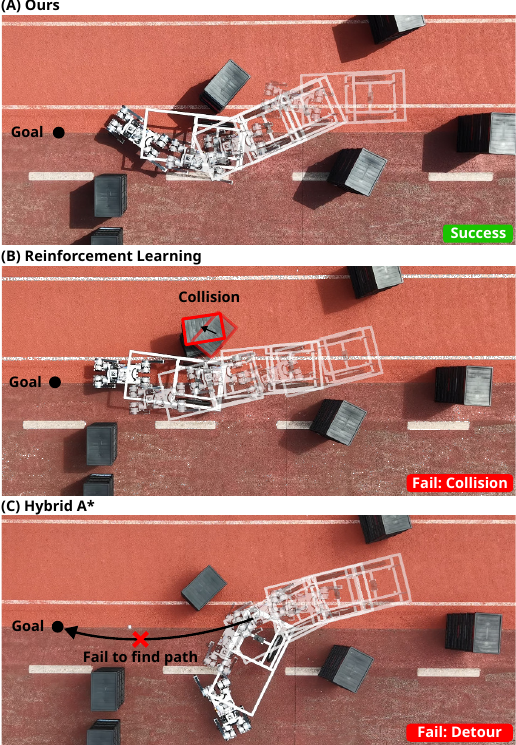}
    \caption{This figure shows a comparison of the RAIBO-Wagon trajectories between our method and the two baselines on Terrain F. (A) shows the trajectory produced by our method, which reaches the goal without collision. (B) shows the trajectory of the RL-based navigation policy, which ends in a collision. (C) shows the trajectory of the Hybrid A* planner, which fails to find a feasible path through the narrow passage.}
    \label{fig:baseline}
\end{figure}

\subsubsection{Baseline Study}

We compare our framework against two baselines that can also be applied to the wagon-towing task.

\textit{Hybrid A*.}
One natural candidate is classical Hybrid A*. Two approaches are possible for this baseline: (i) plan a path with Hybrid A* and track it with a separate path follower, or (ii) execute the motion primitives produced by the planner directly. The first approach requires a separate, wagon-specific path follower, whose tracking errors would be indistinguishable from the planner's own errors. We therefore evaluate only the second, which executes the planner's output directly and isolates its performance as a planner. We search over forward velocities of $\{0.5, 0.6, 0.7, 0.8, 0.9\}$\,m/s and yaw rates of $\{-0.6, -0.3, 0.0, 0.3, 0.6\}$\,rad/s, with a primitive duration of $0.5$\,s (2\,Hz). The map resolution is set to $10$\,cm, matching that of our system. The planner uses a standard rigid-hitch trailer kinematics model, which does not capture the full dynamics of a quadruped robot towing a wagon. We initially considered planning end-to-end from the start to the 15\,m goal, but this was infeasible within the real-time budget. We therefore replan once per cycle from the current robot position to a waypoint 5\,m ahead, a distance tuned to maximize the baseline's performance. The waypoint comes from the same A* planner as in our system.

\begin{table}[t]
\renewcommand{\arraystretch}{1.25}
\caption{Success rate for the baseline study.}
\label{tab:baseline}
\centering
\begin{tabular}{l|ccc}
\hline\hline
 & \textbf{Ours} & \textbf{Hybrid A*} & \textbf{RL} \\
\hline
Success & 5/5 & 0/5 & 0/5 \\
\hline\hline
\end{tabular}
\end{table}

\textit{Reinforcement Learning.}
Alternatively, one may train a waypoint-following policy via reinforcement learning (RL). To construct this baseline, we train the policy with PPO in the RaiSim physics simulator under the same obstacle and wagon-inertia randomization used for CFM model training. The policy is trained to navigate RAIBO-Wagon toward a waypoint placed in a random direction at a
distance sampled between 3 and 5\,m from the robot's current position. At deployment, the A* planner used in our system generates a waypoint 5\,m ahead. Once the robot reaches it, A* replans from the new position until the goal is reached.

\begin{figure*}[t]
    \centering
    \includegraphics[width=0.9\linewidth]{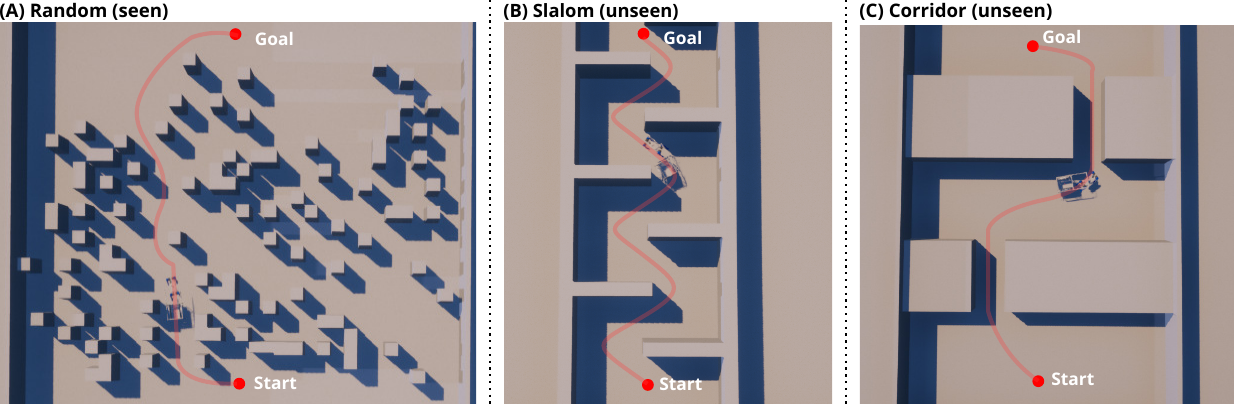}
    \caption{This figure shows the simulation environment setup. (A), Random, is a terrain with randomly placed boxes; it is the environment used to train the CFM model and the FDM. (B) and (C) are custom-designed terrains for evaluating the navigation framework, and neither is seen during training by either model. (B), \textit{Slalom}, evaluates agile turning, while (C), \textit{Corridor}, evaluates precise local navigation through narrow passages.}
    \label{fig:setting}
\end{figure*}

\textit{Results and Analysis.}
As shown in Table~\ref{tab:baseline}, both baselines achieve zero success on Terrain F, whereas our framework succeeds on every trial. Both fail at the same location---the narrowest corridor in Terrain F---but for different reasons, as shown in Fig.~\ref{fig:baseline}. Hybrid A* fails to find a feasible path through the narrow passage and detours along an alternative route. Three factors compound to produce this failure: (i) Hybrid A* is bound to the A*-generated waypoint, which is not guaranteed to be optimal, whereas our method does not strictly follow it; (ii) the 10\,cm map resolution is too coarse for precise planning in tight spaces; and (iii) rigid-hitch trailer kinematics cannot capture the full coupled dynamics of RAIBO-Wagon, restricting the reachable search space. All three limitations are consequences of the real-time budget: given sufficient computation, each could in principle be resolved. This confirms that Hybrid A* is ill-suited for real-time navigation of the RAIBO-Wagon system. By contrast, the RL policy fails for a structurally different reason. Optimizing expected cumulative reward does account for future states, but only implicitly, through a value function that must generalize across the entire training distribution. Tightly constrained configurations such as the narrowest corridor are rare in that distribution, so the learned value is least reliable exactly where precision matters most. This is particularly problematic for wagon navigation, where a locally suboptimal action can place the trailing wagon in a configuration from which collision is unavoidable regardless of subsequent actions. Avoiding such states requires lookahead that is accurate for the configuration at hand, which a reactive policy cannot provide. In summary, our framework outperforms Hybrid A* through the CFM model warm-start, which enables fast convergence to high-quality trajectories within the real-time budget. It outperforms the RL policy through MPPI, which performs this lookahead online by rolling out an explicit dynamics model.

\begin{table*}[t]
\renewcommand{\arraystretch}{1.25}
\caption{Success rate [\%] and navigation time [s] comparison between vanilla MPPI and our method (Flow + MPPI) across three terrain types and three difficulty levels over $N=200$ trials.}
\label{tab:comparison}
\centering
\begin{tabular*}{\textwidth}{@{\extracolsep{\fill}}l|ccc|ccc|ccc}
\hline\hline
& \multicolumn{3}{c|}{Random} & \multicolumn{3}{c|}{Slalom} & \multicolumn{3}{c}{Corridor} \\
Method & Easy & Med & Hard & Easy & Med & Hard & Easy & Med & Hard \\
\hline
\multicolumn{10}{l}{\textit{Success Rate [\%]}} \\
\hline
MPPI (vanilla)   & 81.0\,{\scriptsize($\pm$2.8)} & 68.5\,{\scriptsize($\pm$3.3)} & 61.5\,{\scriptsize($\pm$3.4)} & 89.0\,{\scriptsize($\pm$2.2)} & 66.5\,{\scriptsize($\pm$3.3)} &
45.5\,{\scriptsize($\pm$3.5)} & 90.0\,{\scriptsize($\pm$2.1)} & 75.5\,{\scriptsize($\pm$3.0)} & 29.5\,{\scriptsize($\pm$3.2)} \\
Ours  & \textbf{89.5}\,{\scriptsize($\pm$2.2)} & \textbf{78.5}\,{\scriptsize($\pm$2.9)} & \textbf{73.0}\,{\scriptsize($\pm$3.1)} & \textbf{99.5}\,{\scriptsize($\pm$0.5)} &
\textbf{99.5}\,{\scriptsize($\pm$0.5)} & \textbf{96.5}\,{\scriptsize($\pm$1.3)} & \textbf{98.5}\,{\scriptsize($\pm$0.9)} & \textbf{92.5}\,{\scriptsize($\pm$1.9)} &
\textbf{72.5}\,{\scriptsize($\pm$3.2)} \\
\hline
\multicolumn{10}{l}{\textit{Time [s]}} \\
\hline
MPPI (vanilla)   & 19.3 & 20.4 & 21.0 & 21.9 & 22.5 & 25.1 & 21.8 & 23.2 & 26.4 \\
Ours  & 21.0 & 23.0 & 26.1 & 21.4 & 21.5 & 21.5 & 21.4 & 21.5 &
21.5 \\
\hline\hline
\end{tabular*}
\end{table*}

\subsection{Ablation Study}

\subsubsection{Simulation Evaluation Setup}

\begin{figure}[t]
    \centering
    \includegraphics[width=1.0\linewidth]{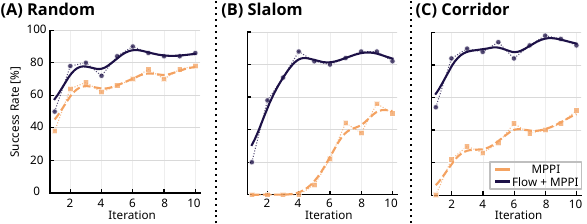}
    \caption{This figure shows the results of the first ablation, vanilla MPPI versus CFM warm start. (A), (B), and (C) show the success rate with respect to the number of MPPI iterations on the Random, Slalom, and Corridor terrains, respectively. The purple solid line (Flow + MPPI) shows the success rate of our framework, which is warm-started from the CFM model, and the orange dashed line (MPPI) shows that of the vanilla MPPI baseline, which is initialized from the previous command.}
    \label{fig:ablation1} 
\end{figure}

We evaluate our framework in three simulation environments, shown in Fig.~\ref{fig:setting}. Random (Fig.~\ref{fig:setting}(A)) is a seen environment, used to train the CFM model and the FDM. Its obstacles are distributed randomly across the terrain. The remaining two are unseen environments designed to stress-test specific navigation capabilities. Slalom (Fig.~\ref{fig:setting}(B)) evaluates agile turning through a sequence of closely spaced obstacles, and Corridor (Fig.~\ref{fig:setting}(C)) evaluates precise local navigation through narrow passages. Difficulty is controlled by a different parameter in each environment---obstacle density in Random, inter-obstacle spacing in Slalom, and corridor spacing in Corridor.

\subsubsection{Vanilla MPPI vs. CFM Warm-Start}

The primary contribution of this work is the integration of a CFM model warm-start into MPPI for navigation while towing a wagon. Here, we conduct an ablation study that isolates the effect of CFM model initialization on MPPI optimization performance.

All hyperparameters, including the number of samples, the number of optimization iterations, and the Gaussian noise schedule, are identical in both conditions. The two differ only in their initialization strategy. The vanilla MPPI baseline initializes from the previous action sequence shifted by one step with additive Gaussian noise, whereas our framework initializes from the CFM model output. Both are evaluated over $N=200$ trials on each terrain of Fig.~\ref{fig:setting}.

\begin{figure*}[t]
    \centering
    \includegraphics[width=1.0\linewidth]{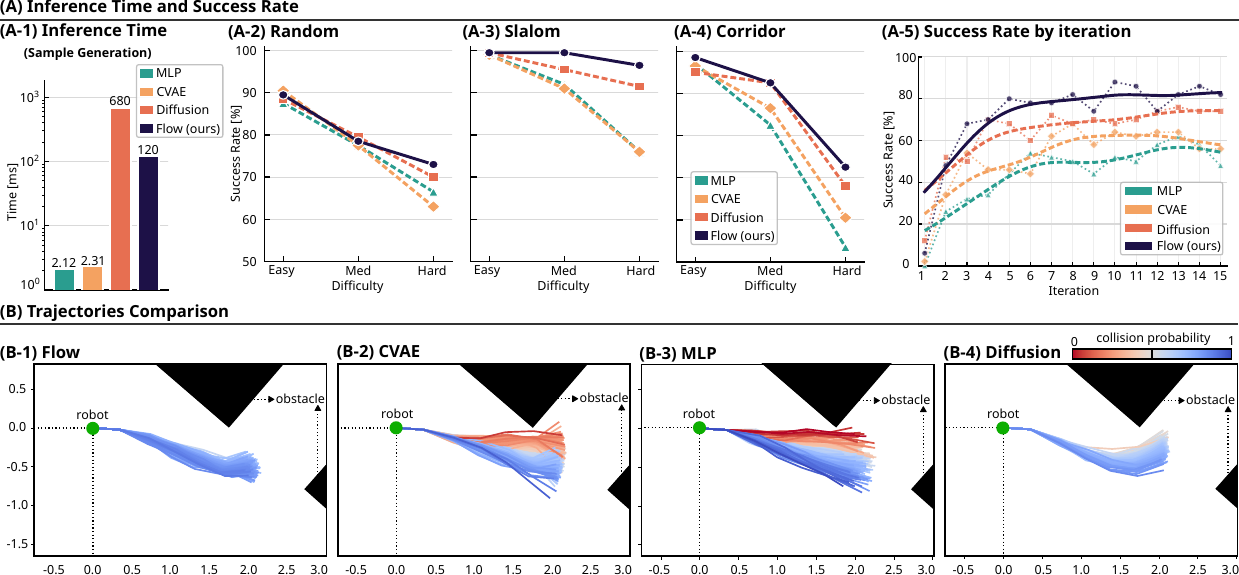}
    \caption{This figure shows the results of the second ablation, warm-start model architecture selection. (A) shows the inference time and success rate of each architecture. (A-1) shows, on a log scale, the inference time for generating the initial samples only, not for the whole planning cycle. Panels (A-2) through (A-4) show the success rate of each architecture with respect to terrain difficulty on Random, Slalom, and Corridor, respectively. (A-5) shows the success rate with respect to the number of MPPI iterations on Slalom at a fixed difficulty. (B) compares the initial samples generated by each architecture in the same scene. The green dot is the current robot position, and the black shapes are obstacles. Each line is colored by its collision probability: blue lines indicate a low collision probability and red lines indicate a high collision probability. Diffusion and Flow denote the DDPM (100 denoising steps) and our CFM model (10 Euler steps), respectively.}
    \label{fig:trajectorycomparison3}
\end{figure*}

Table~\ref{tab:comparison} reports the success rate and total navigation time for each terrain. Our framework consistently outperforms vanilla MPPI across all terrain types, and the gap widens with difficulty. On the hard Slalom and Corridor settings, it achieves more than twice the success rate of the baseline, whereas on easier settings the improvement is smaller. In other words, the benefit of CFM model warm-starting scales with the difficulty of the environment. Table~\ref{tab:comparison} also reports the total navigation time. On the Slalom and Corridor terrains, our framework navigates faster than vanilla MPPI. We attribute this to our framework's ability to find better command sequences in confined spaces, owing to its fast and accurate convergence. On the Random terrain, however, our framework takes longer than vanilla MPPI. This is not due to poor planning performance. Our framework completes more Random episodes than vanilla MPPI, and these additional successes, which vanilla MPPI fails to complete, tend to take longer and thus raise the average time.

The performance gain stems from two complementary properties of the warm-start, each inherited from a different stage of the expert pipeline. First, the CFM model's samples carry global, collision-aware context from the expert path optimizer, which refines an A* path using the learned risk predictor. A shifted copy of the previous action sequence does not encode this context. The two initializations are therefore qualitatively different from the outset, and our framework already starts in a better region of command space at iteration~1. Second, because the CFM model is distilled from the expert's MPPI-optimized solutions rather than from raw trajectories, each sample already embodies the expert optimizer's refinement, giving MPPI a head start.

Fig.~\ref{fig:ablation1} illustrates the effect of the CFM model warm-start by plotting the per-iteration success rate, evaluated over 50 trials per terrain. At iteration~1, after a single MPPI update, the success rate reflects the quality of the initialization alone. Our framework already achieves higher success rates than vanilla MPPI on every terrain at this point, indicating that the CFM model provides a better initialization. Our framework then converges within four iterations on all terrains, whereas vanilla MPPI does not converge within ten. In the Slalom environment in particular (Fig.~\ref{fig:ablation1}(B)), vanilla MPPI achieves 0\% success through the first four iterations. This pattern suggests that in confined spaces the set of feasible command sequences forms a narrow basin that a shifted previous solution rarely reaches, whereas a learned prior places samples inside it from the start.

\subsubsection{Warm-Start Architecture Selection}

A natural question following the previous ablation is whether the CFM model is the best architectural choice for warm-starting MPPI, or whether alternative generative architectures could serve equally well. Here, we investigate this question through a systematic comparison of four candidate architectures: an MLP with Gaussian noise, a Conditional Variational Autoencoder (CVAE)~\cite{kingma2013auto, sohn2015learning}, a Denoising Diffusion Probabilistic Model (DDPM)~\cite{ho2020denoising}, and our Conditional Flow Matching (CFM) model. To ensure a fair comparison, all models have approximately 650K parameters, matching the parameter count of our CFM model. DDPM is evaluated with 100 denoising steps.

We first compare the inference time. Fig.~\ref{fig:trajectorycomparison3}(A-1) reports the mean time for generating the initial samples, measured on the NUC 15 Pro onboard computer, which is used in our real-world experiments. With 100 denoising steps, DDPM is approximately 5.6$\times$ slower than our CFM model, which employs 10 Euler steps.

Fig.~\ref{fig:trajectorycomparison3}(A-2) through (A-4) summarize the success rates across all terrain types. On easy and medium terrains, all four architectures achieve comparable performance, suggesting that the warm-start quality is not a bottleneck when the navigation problem is less geometrically constrained. However, on hard terrains---both seen and unseen---a clear stratification emerges: the CFM model achieves the highest success rate, followed closely by DDPM, while MLP and CVAE trail significantly behind.

\begin{figure*}[t]
    \centering
    \includegraphics[width=1.0\linewidth]{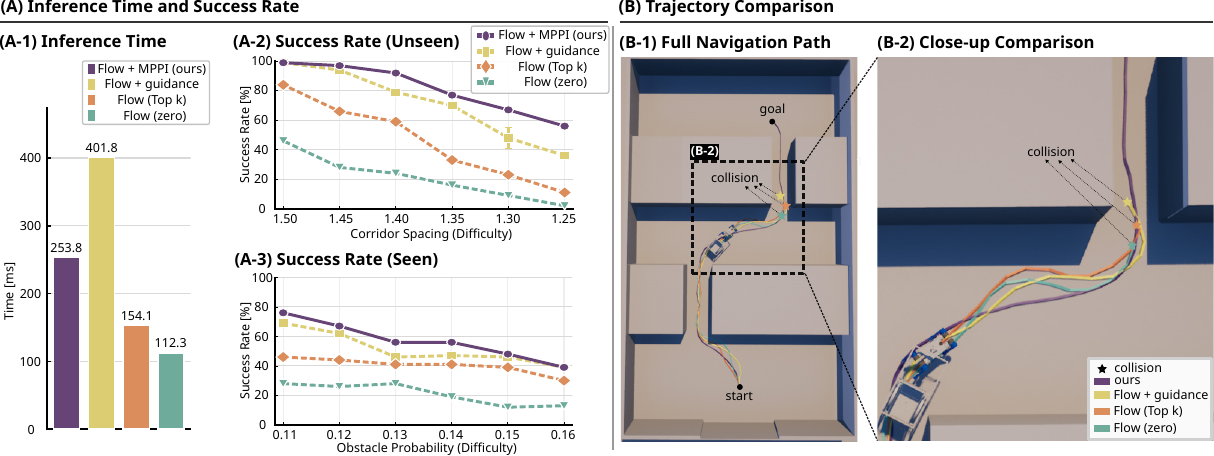}
    \caption{This figure shows the results of the third ablation, alternatives to MPPI refinement. (A) shows the inference time and success rate of each method. (A-1) shows the end-to-end inference time of each method, measured until the final optimized command is produced. (A-2) and (A-3) show the success rate with respect to terrain difficulty on the unseen Corridor terrain and the seen Random terrain, respectively. (B) compares the trajectories of the four methods on the Corridor terrain, where stars mark collisions. (B-1) shows the full navigation trajectory of the RAIBO-Wagon system in simulation, and (B-2) shows a close-up of the region marked by the dashed box in (B-1). Flow (zero), Flow (Top k), and Flow + guidance correspond to zero-start sampling, best-of-$k$ selection, and FDM-guided flow, respectively.}
    \label{fig:ablation3}
\end{figure*}

We found that this performance difference results from the initial sample quality of each architecture. Fig.~\ref{fig:trajectorycomparison3}(B) visualizes representative initial trajectory samples from each architecture under an identical terrain scenario. MLP and CVAE produce samples that are distributed as broad Gaussian clouds around the conditioning input, resulting in a non-negligible fraction of collision-inducing trajectories. In contrast, the CFM model generates tightly structured samples that conform to the terrain geometry, with no collision-inducing trajectories observed. DDPM produces similarly structured samples, though a small number of low-quality trajectories remain. This initial sample distinction arises because flow and diffusion models capture complex, multimodal distributions, whereas the MLP imposes a unimodal Gaussian and the CVAE tends to produce blurred, mode-averaged samples. This distinction in initial sample quality drives the success rate difference.

One might expect that MPPI refinement would compensate for differences in initial sample quality, rendering the choice of warm-start architecture inconsequential. Fig.~\ref{fig:trajectorycomparison3}(A-5) reveals why this intuition does not hold. The figure plots the per-iteration success rate for all four architectures. At iteration 1, after a single MPPI update, the success rates are similar across all methods. As the number of iterations increases, however, the gap widens monotonically.

This behavior reflects a property of MPPI. Since it explores command space through local Gaussian perturbations, the quality of the initial distribution determines the ceiling of what refinement can achieve. CFM model samples already reside in high-reward regions of command space, so each MPPI iteration refines from a strong foundation and converges to superior solutions. In contrast, MLP and CVAE samples are scattered across low-reward regions. Gaussian perturbation cannot fully compensate for such a poor starting point within the iteration budget, so the optimization ceiling is lower. In summary, the CFM model gives MPPI a higher performance ceiling than the other three architectures.

\subsubsection{Alternatives to MPPI Refinement}

The previous ablations established that CFM model warm-starting improves MPPI performance. Our framework also uses a learned forward dynamics model to evaluate command samples. A natural follow-up question is whether some other way of combining the CFM model with the FDM could eliminate the MPPI refinement stage altogether. We investigate this by comparing our framework against three flow-based alternatives that forgo MPPI refinement:

\begin{enumerate}
    \item \textbf{Zero-start sampling}: A single trajectory is generated from the CFM model by integrating from $x_0 = \mathbf{0}$ and executed directly, without FDM-based refinement.
    \item \textbf{Best-of-$k$ selection}: $k = 256$ trajectories are sampled from the CFM model, and the highest-reward one under the FDM is selected.
    \item \textbf{FDM-guided flow}: The flow's vector field is steered during ODE integration by a reward gradient computed from the FDM.
\end{enumerate}

For a fair comparison, all methods use the same CFM model and the same reward function, with the number of flow steps fixed at 10. Fig.~\ref{fig:ablation3}(A-2) and (A-3) report success rates over 100 trials per terrain, stratified by difficulty and terrain novelty. On seen terrains, all four methods perform comparably across difficulty levels. On unseen terrains, however, the performance gap widens markedly with increasing difficulty, and our framework consistently outperforms all three alternatives. Real-time viability depends on cost as much as on accuracy, and the alternatives differ sharply in this respect (Fig.~\ref{fig:ablation3}(A-1)). Zero-start sampling and best-of-$k$ selection are lightweight, requiring only forward passes through the CFM model and, for best-of-$k$, through the FDM. FDM-guided flow, in contrast, backpropagates a reward gradient through the FDM at every integration step. This makes it considerably more expensive than our framework, while also achieving lower success rates.

Fig.~\ref{fig:ablation3}(B) provides a geometric interpretation of this result. The three alternatives follow similar paths, suggesting that none departs far from the CFM model's learned distribution. Our framework, by contrast, deviates from these paths and is the only method that avoids collision. This is consistent with the success-rate ranking on unseen terrains and points to a mechanism. The alternatives all draw their candidates from the CFM model's distribution and consequently struggle when the required trajectory lies away from it. MPPI instead searches outside that distribution through iterative Gaussian perturbations, allowing it to recover solutions that the CFM model alone does not readily produce.

Taken together, the three ablation studies reveal a complementary relationship between the CFM model and MPPI. The CFM model raises the performance MPPI attains within the real-time budget by providing a high-quality initialization concentrated in rewarding regions of command space. Conversely, MPPI searches beyond the CFM model's learned distribution through iterative Gaussian exploration, extending the reachable solution space beyond what imitation alone provides. Each component thus compensates for the main limitation of the other. The CFM model is bound to what it has learned, and MPPI, given a poor starting point, cannot reach high-quality solutions within the time available.

\section{Conclusion and Future Work}
In this work, we introduce a real-time navigation system for a quadruped robot towing a wagon. To handle the complex dynamics introduced by the wagon, we use a CFM model to warm-start an MPPI planner built on a learned forward dynamics model. This combination mitigates two complementary weaknesses: the sensitivity of MPPI to its initial sample distribution, and the tendency of the CFM model to remain confined to its training distribution. We also design a wagon and a wagon-specific locomotion controller, both of which contribute to the overall system. Through extensive real-world experiments, we demonstrate that our system navigates various obstacle-rich terrains in real time, succeeding in 34 of 40 trials across eight configurations and towing a total load of up to 40\,kg.

Despite these results, several limitations remain. Our system is not well suited to tasks that require the robot to maneuver back and forth in confined spaces to align the wagon. Extending the framework to plan such multi-maneuver motions is therefore a promising direction for future work. Additionally, our system focuses on obstacle avoidance. Ground terrain, however, also plays a critical role for the RAIBO-Wagon system, since the terrain traversable by the legged robot differs from that traversable by the wagon. Expanding the navigation framework to incorporate terrain-level reasoning is thus another natural extension. Finally, the CFM model and the FDM are trained for one specific custom-designed wagon. For broader real-world applicability, these models should generalize across wagons of varying shape and size. Developing such a wagon-agnostic navigation system remains an important avenue for future research.

\appendices
\section{Low-Level Controller Validation}
\label{app:lowlevel}

\begin{figure}[t]
    \centering
    \includegraphics[width=0.9\linewidth]{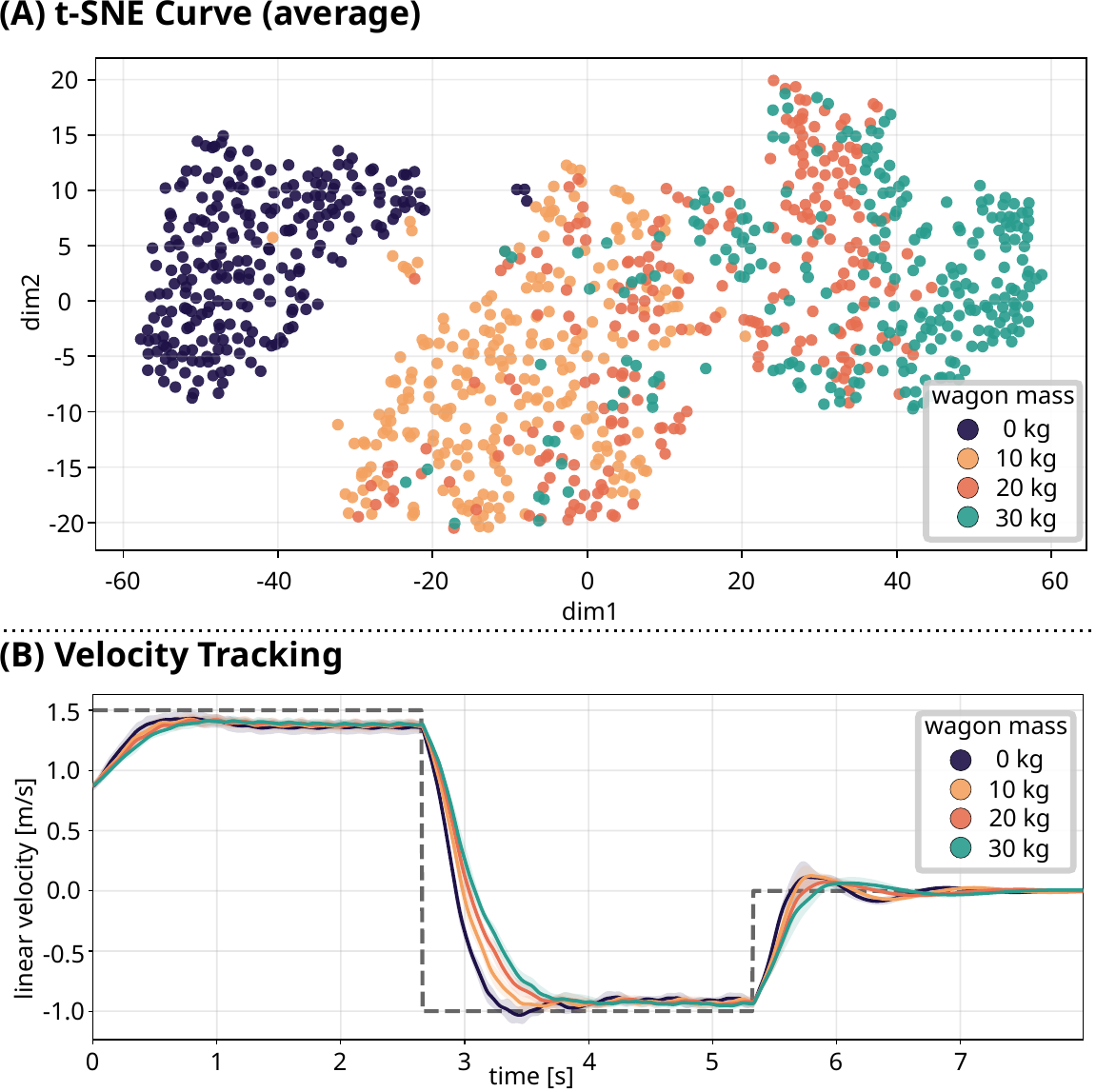}
    \caption{(A) shows the t-SNE visualization of the GRU encoder latent vectors, averaged over one locomotion cycle, across varying wagon masses. (B) shows the command tracking performance across different wagon masses, where the dashed line indicates the commanded velocity and the four solid lines represent the measured robot body linear velocity under each configuration.}
    \label{fig:lowlevelcurve}
\end{figure}

In this paper, we introduce a locomotion controller designed specifically for wagon towing. As described in Sec.~\ref{sec:lowmethod}, existing locomotion controllers are inadequate for this task, as they do not account for wagon inertia during locomotion. Here, we validate that our wagon-specific controller implicitly encodes wagon mass and achieves reliable command velocity tracking across a range of wagon configurations.

Fig.~\ref{fig:lowlevelcurve}(A) shows a t-SNE visualization of the latent space produced by the locomotion controller's GRU encoder (Fig.~\ref{fig:new_method}). To isolate the effect of wagon mass, we average the latent vectors over one locomotion cycle and visualize their distribution across varying wagon masses. The no-wagon case (0\,kg) and the loaded case (30\,kg) occupy clearly separated regions, and intermediate masses (between 0 and 30\,kg) form a gradual transition between them. This indicates that the controller implicitly encodes wagon mass from proprioceptive feedback alone, without receiving any wagon state as input.

Fig.~\ref{fig:lowlevelcurve}(B) shows forward velocity tracking performance across different wagon masses. We focus on forward velocity because wagon mass primarily degrades forward locomotion, whereas angular velocity tracking remains largely unaffected owing to the passive joint in our wagon design. The controller accurately tracks the commanded velocity across all wagon-mass configurations. The main difference is in transient response: lighter configurations respond faster to abrupt command changes (e.g., a step from $+1.5$ to $-1.0$\,m/s), while heavier configurations respond more gradually because of the increased
inertia.

Together, these results confirm that our wagon-specific controller implicitly accounts for wagon mass and maintains reliable command tracking across the full range of tested configurations.

\section{FDM Validation}
\label{app:fdm}
In this paper, we employ a learned forward dynamics model (FDM) to evaluate the sampled trajectories, which makes our sampling-based method feasible within the real-time budget. In principle, other choices are possible: a simple kinematic model of the robot and the wagon, or a full dynamics model that relies on a high-fidelity rigid-body simulator such as RaiSim. Here, we validate our choice of the learned FDM over these alternatives and show why it is essential for our task.

For a fair comparison, we run MPPI with the kinematic model using the same number of samples and iterations. For the RaiSim-based full dynamics model, however, matching the number of samples and iterations is infeasible. Under this setting, the optimization takes about 5 minutes on the onboard computer to generate a 0.5\,s command, far beyond any real-time budget. We therefore evaluate the full dynamics model with a single iteration and 16 samples. While this configuration is still impractical for real-time onboard optimization, it is at least tractable enough to serve as a reference.

\begin{figure}[t]
    \centering
    \includegraphics[width=1.0\linewidth]{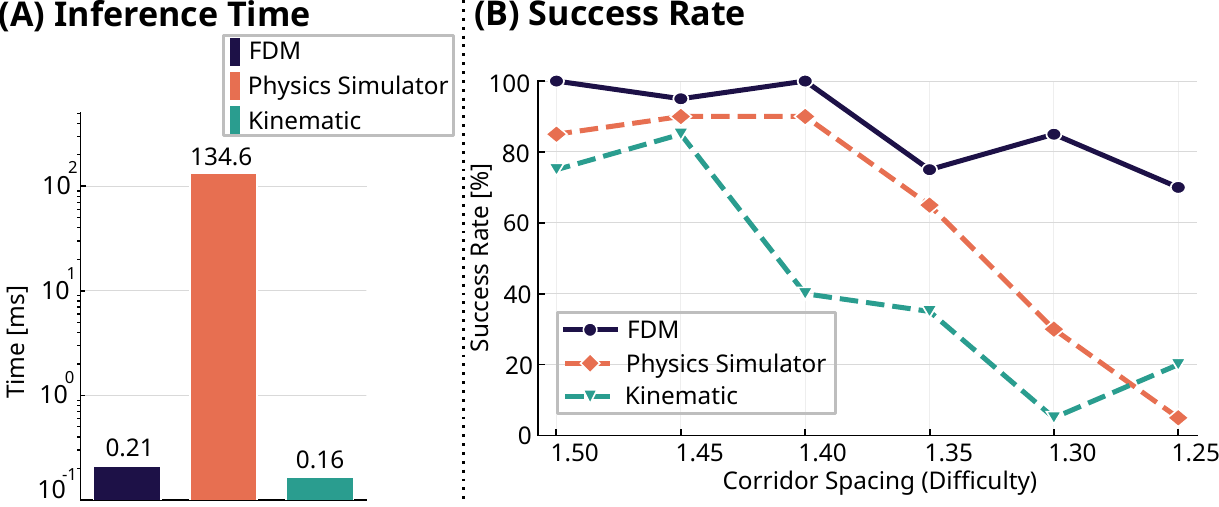}
    \caption{(A) shows the inference time of each dynamics model, measured for a single rollout sample. (B) shows the success rate obtained with each dynamics model.}
    \label{fig:dynamics_model}
\end{figure}

We evaluate the effect of the dynamics model on the Corridor terrain (Fig.~\ref{fig:setting}(C)). As shown in Fig.~\ref{fig:dynamics_model}(B), the kinematic model degrades with difficulty, as it cannot capture the coupled quadruped-wagon dynamics. Even though the full dynamics model uses the same simulator for both rollout and evaluation, its success rate drops sharply as terrain difficulty increases, whereas our learned FDM remains robust. This suggests that performing many refinement iterations with a sufficiently large sample size is essential for our task---something the full dynamics model cannot afford within the time budget. Fig.~\ref{fig:dynamics_model}(A) reports the corresponding inference time: the full dynamics model is more than 100$\times$ slower than the FDM, while the FDM and the kinematic model are comparable. Taken together, these results show that a dynamics model that is both accurate and fast to evaluate is essential for finding feasible solutions in narrow spaces under a tight time budget.

\vspace{11pt}

\bibliographystyle{IEEEtran}
\bibliography{references} 

\end{document}